\documentclass[sigconf, screen]{acmart}

\usepackage{multirow}
\usepackage{makecell}
\usepackage{tabularx}
\usepackage{placeins}
\usepackage{float}
\usepackage{dblfloatfix}
\usepackage{placeins}
\usepackage{xurl}
\usepackage{float}
\newcolumntype{Y}{>{\raggedright\arraybackslash}X}

\renewcommand\footnotetextcopyrightpermission[1]{}
\acmConference[MM '26]{Proceedings of the 34th ACM International Conference on Multimedia}{10--14 November 2026}{Rio de Janeiro, Brazil}

\AtBeginDocument{%
  }

\begin{document}

\title[GAG: A Plug-and-Play Framework for Private Knowledge Injection in Large Language Models]
{Generation-Augmented Generation: A Plug-and-Play Framework for Private Knowledge Injection in Large Language Models}


\author{Rongji Li}
\affiliation{%
  \institution{MAIS, Institute of Automation, Chinese Academy of Sciences}
  \city{Beijing}
  \country{China}
}
\affiliation{%
  \institution{School of Artificial Intelligence, University of Chinese Academy of Sciences}
  \city{Beijing}
  \country{China}
}
\affiliation{%
  \institution{Zhongguancun Academy}
  \city{Beijing}
  \country{China}
}
\email{lirongji2025@ia.ac.cn}

\author{Jian Xu}
\authornote{Corresponding author.}
\affiliation{%
  \institution{MAIS, Institute of Automation, Chinese Academy of Sciences}
  \city{Beijing}
  \country{China}
}
\affiliation{%
  \institution{School of Artificial Intelligence, University of Chinese Academy of Sciences}
  \city{Beijing}
  \country{China}
}
\affiliation{%
  \institution{Zhongguancun Academy}
  \city{Beijing}
  \country{China}
}
\email{jian.xu@ia.ac.cn}

\author{Yi Chen}
\affiliation{%
  \institution{MAIS, Institute of Automation, Chinese Academy of Sciences}
  \city{Beijing}
  \country{China}
}
\affiliation{%
  \institution{School of Artificial Intelligence, University of Chinese Academy of Sciences}
  \city{Beijing}
  \country{China}
}
\affiliation{%
  \institution{Zhongguancun Academy}
  \city{Beijing}
  \country{China}
}
\email{yi.chen@nlpr.ia.ac.cn}

\author{Xueqing Chen}
\affiliation{%
  \institution{Zhongguancun Academy}
  \city{Beijing}
  \country{China}
}
\email{xqchen@cnic.cn}

\author{Yisheng Yang}
\affiliation{%
  \institution{MAIS, Institute of Automation, Chinese Academy of Sciences}
  \city{Beijing}
  \country{China}
}
\affiliation{%
  \institution{School of Advanced Interdisciplinary Sciences, University of Chinese Academy of Sciences}
  \city{Beijing}
  \country{China}
}
\email{yangyisheng2025@ia.ac.cn}

\author{Jiayi Wang}
\affiliation{%
  \institution{MAIS, Institute of Automation, Chinese Academy of Sciences}
  \city{Beijing}
  \country{China}
}
\affiliation{%
  \institution{School of Advanced Interdisciplinary Sciences, University of Chinese Academy of Sciences}
  \city{Beijing}
  \country{China}
}
\email{wangjiayi2025@ia.ac.cn}

\author{Xingyu Chen}
\affiliation{%
  \institution{Zhongguancun Academy}
  \city{Beijing}
  \country{China}
}
\email{chenxy.sean@gmail.com}

\author{Chunyu Xie}
\affiliation{%
  \institution{360 AI Research}
  \city{Beijing}
  \country{China}
}
\email{xiechunyu@360.cn}

\author{Dawei Leng}
\authornotemark[1]
\affiliation{%
  \institution{360 AI Research}
  \city{Beijing}
  \country{China}
}
\email{lengdawei@360.cn}

\author{Xu-Yao Zhang}
\affiliation{%
  \institution{MAIS, Institute of Automation, Chinese Academy of Sciences}
  \city{Beijing}
  \country{China}
}
\affiliation{%
  \institution{School of Artificial Intelligence, University of Chinese Academy of Sciences}
  \city{Beijing}
  \country{China}
}
\affiliation{%
  \institution{Zhongguancun Academy}
  \city{Beijing}
  \country{China}
}
\email{xyz@nlpr.ia.ac.cn}

\renewcommand{\shortauthors}{Rongji Li et al.}

\begin{abstract}
In domains such as materials science, biomedicine, and finance, high-stakes deployment of large language models (LLMs) requires injecting private, domain-specific knowledge that is proprietary, fast-evolving, and under-represented in public pretraining. However, the two dominant paradigms for private knowledge injection each have clear drawbacks: fine-tuning is expensive to iterate under continual updates that can induce catastrophic forgetting and general-capability regression; retrieval-augmented generation (RAG) keeps the base model intact but remains brittle in specialized private corpora due to chunk-induced evidence fragmentation, retrieval mismatch, and long-context pressure. Inspired by how multimodal LLMs align heterogeneous modalities into a shared semantic space, we propose \textbf{Generation-Augmented Generation (GAG)}, which treats private expertise as an auxiliary modality and injects it into a frozen base model through a compact, constant-budget latent interface. Concretely, GAG distills question-conditioned specialist knowledge from lightweight domain experts into \textbf{multi-slot latent memories}, integrates multi-layer expert signals via \textbf{per-slot cross-layer fusion}, and aligns them to the frozen base model through \textbf{gated residual projection}, while supporting scalable mixed-domain deployment with reliable selective activation. In a unified mixed-domain evaluation spanning two scientific private-domain QA benchmarks (catalytic materials and immunology adjuvant) together with general-domain queries, GAG consistently outperforms strong retrieval-based and parameter-efficient fine-tuning baselines on specialist QA, while preserving general-domain capability, achieving highly reliable routing, and offering a favorable efficiency--effectiveness trade-off. Code and datasets are provided in the supplementary material. Code is publicly available at https://github.com/360CVGroup/GAG.
\end{abstract}



\maketitle

\section{Introduction}

Large language models (LLMs) have demonstrated strong capabilities across a wide range of natural language processing tasks, including text understanding, generation, and instruction following \citep{grattafiori2024llama,yang2025qwen3,liu2024deepseek,guo2025deepseek}. 
Pretrained on vast corpora of general text, LLMs have profoundly impacted various aspects of daily life and professional environments. However, despite these impressive general capabilities, enabling LLMs to perform optimally in private domains remains a significant challenge. 
In private-domain deployments such as materials science, biomedicine, and finance \citep{chen2023matchat,bao2023disc,chen2023disc}, reliable performance often requires incorporating domain-specific knowledge beyond open-domain pretraining, where expert terminology and conventions are critical for accurate outputs.

\begin{figure}[t]
\centering
\includegraphics[width=\columnwidth]{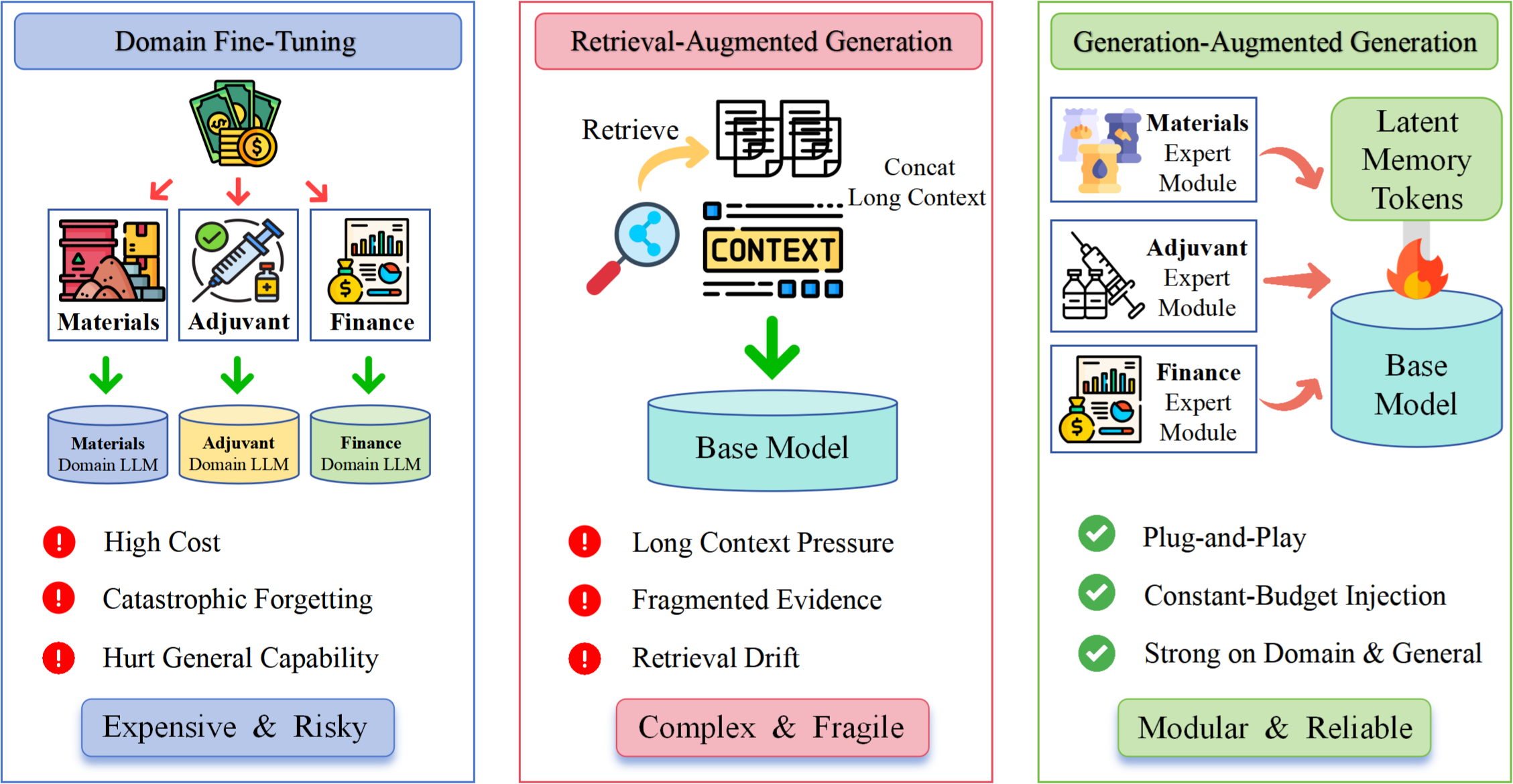}
\caption{\textbf{Three paradigms for private knowledge injection.}
(a) Fine-tuning is expensive and risky.
(b) RAG is complex and fragile due to retrieval and long-context pressure.
(c) \textbf{GAG} treats private-domain knowledge as an auxiliary modality and injects it into a frozen base model through a constant-budget latent interface with selective activation.}
\label{fig:intro-paradigms}
\end{figure}

Two dominant paradigms are commonly used to inject private knowledge into LLMs.
\textbf{(i) Domain fine-tuning} can internalize domain knowledge, but it is costly to iterate, requires careful validation, and risks general-capability regression and catastrophic forgetting under continual updates \citep{gururangan2020don,hu2022lora,dettmers2023qlora}. 
\textbf{(ii) Retrieval-augmented generation (RAG)} preserves the base model by retrieving textual evidence at inference time \citep{lewis2020retrieval,guu2020retrieval,izacard2021leveraging,izacard2023atlas}. 
However, in private domains RAG is often brittle: evidence is fragmented by chunking, retrieval can drift or miss crucial context, and even relevant passages must compete for limited context budget and are unevenly utilized by long-context LLMs \citep{liu2024lost}. 
These limitations suggest that private knowledge injection should be treated as a representation-transfer problem rather than merely text retrieval or repeated parameter updating. Figure~\ref{fig:intro-paradigms} shows these trade-offs and positions \textbf{Generation-Augmented Generation (GAG)} as a constant-budget, modular alternative to both fine-tuning and retrieval-based injection.

In this work, we reformulate private knowledge injection from a multimodal perspective, conceptualizing private-domain knowledge as an auxiliary modality beyond the general-domain language space, rather than as a collection of textual snippets. 
This auxiliary modality can be aligned and fused into a general LLM through lightweight parameter-efficient interfaces, similar in spirit to how multimodal systems connect heterogeneous signals to frozen language backbones \citep{alayrac2022flamingo,li2023blip,liu2023visual,huang2023language}. 
Related design patterns also appear in multimodal systems that align language backbones with video, meteorological, and graph-structured scientific modalities \citep{zhang2023video,zhu2023languagebind,tang2025meteorpred,ock2024multimodal}.
Based on this perspective, we introduce \textbf{GAG}, a retrieval-free and plug-and-play framework that injects private knowledge into a frozen base model through a constant-budget latent interface, without updating the base-model parameters. 
Concretely, GAG adopts a decoupled architecture consisting of a general-purpose base LLM and lightweight domain-specific expert modules. For each domain, a small expert model is first adapted to the target corpus and then specialized to generate question-conditioned background knowledge. Instead of passing such knowledge to the base model as retrieved text, GAG extracts multi-layer hidden states from the expert, compresses them into a small set of memory slots, performs per-slot cross-layer fusion, and projects the resulting latent memories into the embedding space of the frozen base model through a gated residual projector. The injected memories are then consumed through a fixed number of special tokens, transforming private-knowledge injection into a compact representation-level operation. To further support modular multi-domain deployment, GAG incorporates a prototype plug-and-play routing mechanism that selectively activates the general route or domain-specific routes without retraining the backbone, enabling scalable specialist extension while preserving the general-domain capability of the base LLM.

We evaluate GAG in a unified mixed-domain setting covering general-domain queries and two scientific private domains, catalytic materials and immunology adjuvant.  Extensive experiments show that GAG substantially improves specialist-domain performance over strong retrieval-based and parameter-efficient fine-tuning baselines while preserving general-domain capability. In this mixed-domain evaluation, the Prototype Plug-and-Play Routing mechanism achieves highly reliable selective activation, showing that domain-specific expert modules can be attached incrementally in a truly plug-and-play manner. Finally, efficiency analysis further shows that, by replacing retrieved textual evidence with compact latent memory injection, GAG achieves a more favorable efficiency--effectiveness trade-off than our strong RAG baseline, reducing additional token overhead and long-context burden while delivering stronger specialist responses.

\textbf{Contributions.}
(1) We introduce GAG, a retrieval-free and plug-and-play framework that treats private-domain knowledge as an auxiliary modality and injects it into a frozen base model through a constant-budget latent interface. 
(2) We propose a multi-slot latent memory injection design that distills question-conditioned domain knowledge from lightweight expert models into compact multi-layer memories and transfers them to the frozen base LLM via per-slot cross-layer fusion and gated residual projection, together with prototype-based plug-and-play routing for modular multi-domain extension. 
(3) We conduct extensive experiments in a unified mixed-domain setting covering general-domain queries and two scientific private domains, showing that GAG substantially strengthens specialist-domain capability, preserves general-domain performance, enables highly reliable selective routing, and achieves better efficiency than a strong retrieval-based baseline.

\section{Related Work}
\label{sec:related-work}

\subsection{Fine-tuning-based knowledge injection}
Parametric adaptation injects domain knowledge into language models through continued pretraining or supervised fine-tuning on domain-specific data \citep{gururangan2020don}. 
While effective, a central challenge is that continual domain updates may induce catastrophic forgetting and regression of general-domain capability unless additional continual-learning mechanisms are introduced \citep{kirkpatrick2017overcoming,li2017learning}. 
To reduce adaptation cost, parameter-efficient fine-tuning (PEFT) methods update only a small subset of parameters, including adapters, prefix tuning, prompt tuning, and other lightweight modules \citep{houlsby2019parameter,li2021prefix,lester2021power,zaken2022bitfit,liu2022p}. 
Low-rank and quantization-aware variants further improve efficiency and memory usage \citep{hu2022lora,zhang2023adalora,mao2022unipelt,pfeiffer2021adapterfusion,dettmers2023qlora,lialin2023scaling}. 
However, even PEFT still relies on iterative parameter updating and repeated validation, and thus remains less suitable for deployment scenarios that require a strictly frozen base model for governance, stability, and regression control. 
These limitations motivate modular knowledge injection mechanisms that can enhance specialist capability while preserving a reusable frozen backbone.

\subsection{Retrieval-augmented knowledge injection}
Retrieval-augmented generation (RAG) injects external knowledge by retrieving evidence from a corpus and conditioning the language model on retrieved text, and it has become a widely adopted paradigm for knowledge-intensive question answering \citep{lewis2020retrieval,guu2020retrieval,izacard2021leveraging}. 
A large body of work improves retrieve-then-read systems through stronger dense retrieval, late-interaction matching, improved training objectives, and tighter reader-side fusion \citep{karpukhin2020dense,xiong2020approximate,khattab2020colbert,izacard2023atlas,borgeaud2022improving,shi2024replug,khandelwal2019generalization}. 
More recently, language-model-based generation, verification, and self-reflection signals have also been explored to improve retrieval robustness and attribution faithfulness \citep{gao2023precise,gao2023rarr,asai2024self}. 
Despite these advances, RAG remains particularly challenging in private and fast-evolving domains: evidence is fragmented by chunking, top-$k$ retrieval does not guarantee complete coverage, and multiple passages must compete within a finite context budget, where long-context LLMs may under-utilize or misinterpret relevant spans \citep{liu2024lost,bai2024longbench}. 
Context-compression methods reduce prompt overhead, but remain retrieval-dependent and therefore still hinge on retrieval coverage, indexing quality, and evidence selection quality \citep{cheng2024xrag}. 
Auxiliary-model-based transfer methods can inject domain signals through prompt-time mediation \citep{li2025blade}, but they still rely on textual handoff, remain subject to context-budget pressure, and often require cumbersome dataset preparation with high-quality supervision.
In contrast, our work targets retrieval-free private knowledge injection under a frozen base model. Rather than serializing external evidence into text, GAG transfers domain knowledge through a constant-budget latent interface, thereby reducing dependence on retrieval coverage and long-context utilization.

\section{Problem Formulation}
\label{sec:problem-setup}

We consider question answering with a frozen base model.
Let $p_{\theta}(y\mid x)$ denote the conditional distribution induced by a pretrained model with parameters $\theta$, where $x$ is a user query and $y$ is the target answer.
After deployment, $\theta$ is not allowed to be updated.

\textbf{Multi-domain private knowledge.}
Queries are drawn from a mixture of one general-domain distribution $\mathcal{D}_0$ and $N$ private-domain distributions $\{\mathcal{D}_i\}_{i=1}^{N}$.
For each private domain $i$, samples $(x,y)\sim\mathcal{D}_i$ require domain-specific knowledge that is not reliably covered by open-domain pretraining.
We assume each private domain $i$ is associated with a private knowledge source $\mathcal{K}_i$ (e.g., proprietary documents or curated specialist resources), while the general domain $\mathcal{D}_0$ requires no domain-specific augmentation.

\textbf{Knowledge injection as conditional generation with auxiliary side information.}
Our goal is to enable the frozen base model to answer private-domain queries without modifying its parameters.
To this end, we allow the model to condition on an auxiliary injected signal $z$ derived from $(x,\mathcal{K}_i)$:
\begin{equation}
z \;=\; \mathcal{A}_i(x,\mathcal{K}_i), \qquad
\hat{y} \sim p_{\theta}(y \mid x, z),
\label{eq:setup-conditional}
\end{equation}
where $\mathcal{A}_i$ denotes a domain-specific injection mechanism.
Different from retrieval-based formulations that append variable-length textual evidence to the prompt, we are interested in compact latent side information, so that private knowledge can be transferred to the frozen base model through a constant-budget external interface.

\textbf{Objective and constraints.}
Our objective is to improve private-domain QA quality while preserving the general-domain capability of the base model.
Let $\mathcal{L}(\hat{y},y)$ denote a task loss or an evaluation-aligned surrogate.
We seek injection mechanisms $\{\mathcal{A}_i\}_{i=1}^{N}$ such that
\begin{equation}
\begin{aligned}
\min_{\{\mathcal{A}_i\}_{i=1}^{N}} \quad
& \sum_{i=1}^{N}\mathbb{E}_{(x,y)\sim\mathcal{D}_i}\!\left[\mathcal{L}(\hat{y}_i(x),y)\right] \\
\text{s.t.}\quad
& \mathbb{E}_{(x,y)\sim\mathcal{D}_0}\!\left[\mathcal{L}(\hat{y}_0(x),y)\right]
\le R_0 + \epsilon ,
\end{aligned}
\label{eq:setup-objective}
\end{equation}
where $R_0$ denotes the baseline risk of the frozen base model on $\mathcal{D}_0$ without knowledge injection, and $\epsilon$ is an allowable regression margin.

\textbf{Plug-and-play domain expansion.}
We further require modular multi-domain expansion:
when a new domain $k$ arrives, the system should incorporate $\mathcal{A}_k$ without modifying the base-model parameters $\theta$ or previously deployed mechanisms $\{\mathcal{A}_i\}_{i<k}$.
This captures the practical requirement that private knowledge evolves while the base model must remain stable and reusable under incremental specialist extension.

\section{Methodology}
\label{sec:method}

\begin{figure}[t]
\centering
\includegraphics[width=\columnwidth]{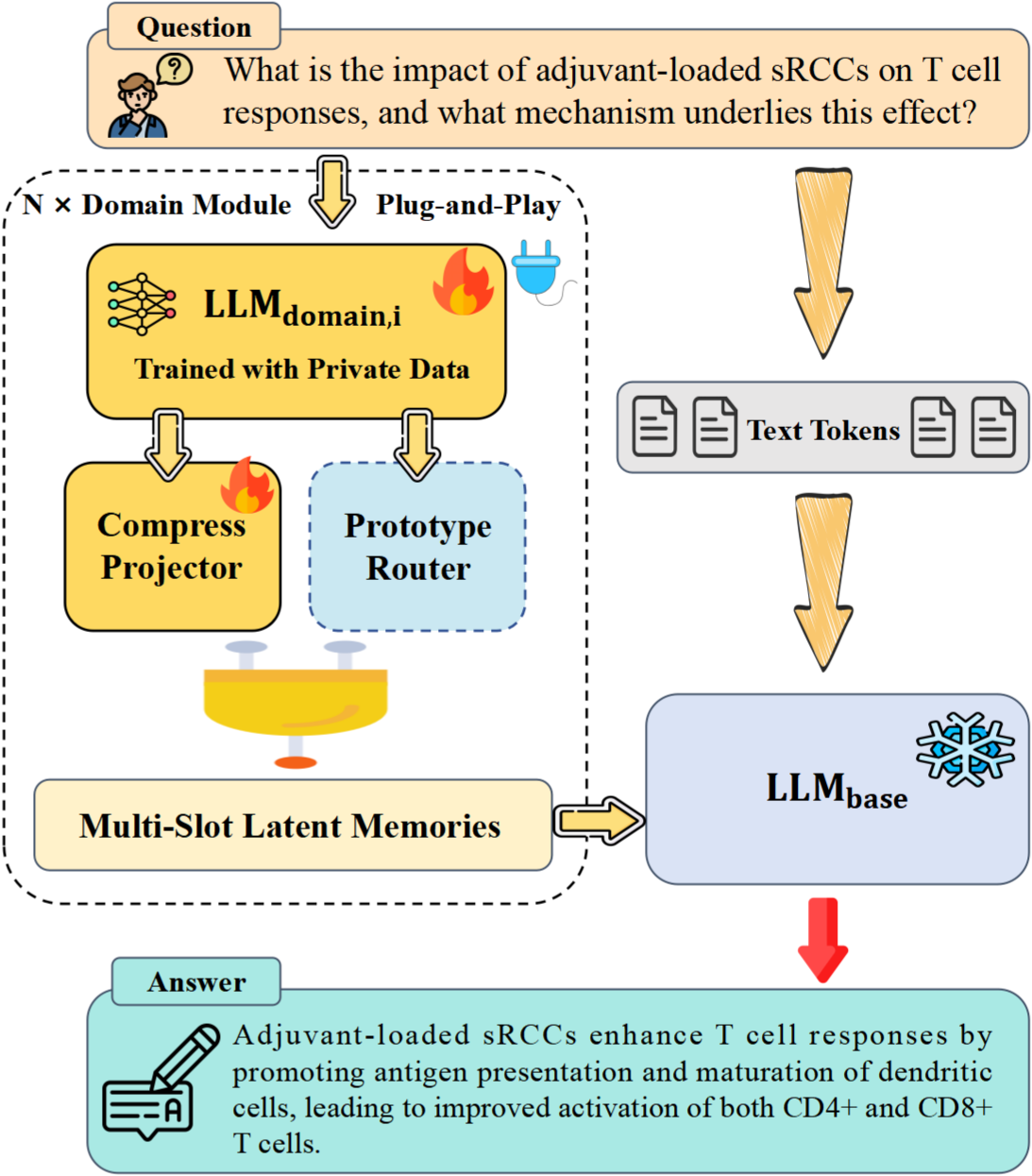}
\caption{\textbf{GAG overview at inference time.} A user query follows the standard textual path into the frozen base model, while routed specialist knowledge is injected in parallel as projected multi-slot latent memories produced by the selected plug-and-play domain module. The router selects between the general path and domain-specific modules without modifying the base model.}
\label{fig:gag-overview}
\end{figure}

\begin{figure*}[t]
\centering
\includegraphics[width=\textwidth]{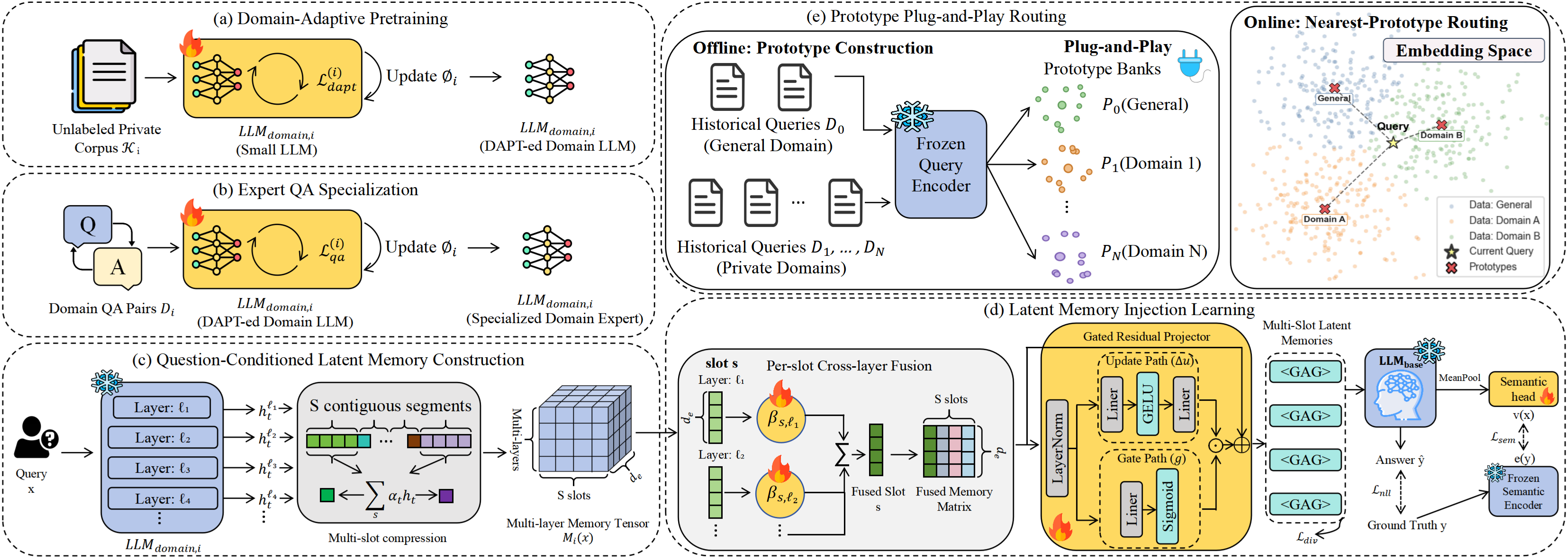}
\caption{\textbf{Detailed methodology of GAG.} (a) Domain-Adaptive Pretraining learns a specialist corpus prior from unlabeled private data. (b) Expert QA Specialization turns the same small model into a query-aware domain expert. (c) The expert’s generated hidden trajectories are compressed into a stabilized multi-layer memory tensor. (d) Injection-side learning performs per-slot cross-layer fusion, gated residual projection, and joint optimization with $\mathcal{L}_{\mathrm{nll}}$, $\mathcal{L}_{\mathrm{sem}}$, and $\mathcal{L}_{\mathrm{div}}$ to align latent memories to the frozen base model. (e) Prototype Plug-and-Play Routing builds prototype banks offline and selects routes online by nearest-prototype matching for training-free incremental deployment.} 
\label{fig:gag-training}
\end{figure*}

In this paper, we present \textbf{Generation-Augmented Generation (GAG)}, a retrieval-free framework for private knowledge injection into a strictly frozen base model. The key idea is to treat private-domain knowledge as an auxiliary modality and transfer it through a compact latent interface rather than retrieved text or repeated backbone adaptation. GAG comprises four components: domain expert construction, question-conditioned latent memory construction, latent memory injection learning, and Prototype Plug-and-Play Routing. Figure~\ref{fig:gag-overview} gives the inference overview, and Figure~\ref{fig:gag-training} details the full methodology.

\subsection{Overview of GAG}
\label{sec:method-overview}

Let $\mathrm{LLM}_{\mathrm{base}}$ denote a frozen base model with parameters $\theta$ and hidden size $d_b$. 
Given a query $x$, GAG either follows the general route or activates a domain-specific route. 
For a selected private domain $i$, the corresponding specialist module produces a fixed-budget latent memory
\begin{equation}
\mathbf{Z}_i(x)=\left[\mathbf{z}_{i,1}(x),\ldots,\mathbf{z}_{i,S}(x)\right]\in\mathbb{R}^{S\times d_b},
\label{eq:gag-overview-memory}
\end{equation}
where $S$ is the number of injected memory slots. 
The base model then generates the answer conditioned on the query and the injected latent memory:
\begin{equation}
\hat{y}\sim p_{\theta}\!\left(y\mid x,\mathbf{Z}_i(x)\right).
\label{eq:gag-overview-gen}
\end{equation}
For the general route, no private-domain memory is injected. 
Unlike retrieval-based approaches, the amount of injected side information is constant with respect to corpus size.

\subsection{Domain Expert Construction}
\label{sec:method-expert}

For each private domain $i$, GAG constructs a lightweight expert model $\mathrm{LLM}_{\mathrm{domain},i}$ with parameters $\phi_i$ and hidden size $d_e$. 
This expert serves as the source of question-conditioned specialist knowledge.

\subsubsection{Stage I: Domain-Adaptive Pretraining}
\label{sec:method-stage1}

As illustrated in Figure~\ref{fig:gag-training}(a), we first adapt $\mathrm{LLM}_{\mathrm{domain},i}$ on the corresponding domain corpus $\mathcal{K}_i$ using standard causal language modeling:
\begin{equation}
\mathcal{L}_{\mathrm{dapt}}^{(i)}
=
\mathbb{E}_{u\sim\mathcal{K}_i}
\left[
-\sum_{t=1}^{|u|}
\log p_{\phi_i}(u_t\mid u_{<t})
\right].
\label{eq:loss-dapt}
\end{equation}
This stage equips the expert with specialist terminology, discourse patterns, and domain-specific regularities.

\subsubsection{Stage II: Expert QA Specialization}
\label{sec:method-stage2}

As illustrated in Figure~\ref{fig:gag-training}(b), after domain-adaptive pretraining, we continue training the expert on in-domain QA pairs $(x,y)\sim\mathcal{D}_i$:
\begin{equation}
\mathcal{L}_{\mathrm{qa}}^{(i)}
=
\mathbb{E}_{(x,y)\sim\mathcal{D}_i}
\left[
-\sum_{t=1}^{|y|}
\log p_{\phi_i}(y_t\mid y_{<t},x)
\right].
\label{eq:loss-qa}
\end{equation}
This stage teaches the expert to activate domain knowledge in a query-aware manner and to produce background information that is directly useful for downstream answering.

Together, Stages I and II yield a specialist expert that is both domain-aware and question-aware (Figure~\ref{fig:gag-training}(a)--(b)): it internalizes private knowledge at the corpus level while also learning how to express that knowledge under task-specific questioning.

\subsection{Question-Conditioned Latent Memory Construction}
\label{sec:method-memory}

As summarized in Figure~\ref{fig:gag-training}(c), given a query $x$, the selected expert first generates a background sequence
\begin{equation}
b_{1:T}\sim p_{\phi_i}(b\mid x),
\label{eq:bg}
\end{equation}
where $b_{1:T}$ is used only as an intermediate carrier and is not exposed to the frozen base model as textual context.

Let $\mathbf{h}^{(\ell)}_t\in\mathbb{R}^{d_e}$ denote the hidden state of the expert at generation step $t$ and layer $\ell$. 
Instead of reading out a single final-layer vector, we collect a set of layer-wise hidden trajectories:
\begin{equation}
\mathbf{H}^{(\ell)}(x)
=
\left[\mathbf{h}^{(\ell)}_1,\ldots,\mathbf{h}^{(\ell)}_T\right]
\in\mathbb{R}^{T\times d_e},
\qquad
\ell\in\mathcal{L},
\label{eq:hidden-traj}
\end{equation}
where $\mathcal{L}$ denotes a selected set of expert layers.

\textbf{Multi-slot compression.}
For each layer $\ell$, we partition the generated sequence into $S$ contiguous segments and compress each segment into one slot. 
Let $\mathcal{T}_s^{(\ell)}$ denote the token indices assigned to slot $s$ at layer $\ell$. 
We compute segment-wise importance weights
\begin{equation}
\alpha_t^{(\ell,s)}
=
\frac{\exp\big(\|\mathbf{h}^{(\ell)}_t\|_2/\tau\big)}
{\sum_{u\in\mathcal{T}_s^{(\ell)}}\exp\big(\|\mathbf{h}^{(\ell)}_u\|_2/\tau\big)},
\qquad
t\in\mathcal{T}_s^{(\ell)},
\label{eq:slot-weight}
\end{equation}
and obtain the slot representation
\begin{equation}
\mathbf{m}^{(\ell)}_{i,s}(x)
=
\sum_{t\in\mathcal{T}_s^{(\ell)}}
\alpha_t^{(\ell,s)}\,\mathbf{h}^{(\ell)}_t
\in\mathbb{R}^{d_e}.
\label{eq:slot-memory}
\end{equation}
This yields a structured multi-layer memory tensor
\begin{equation}
\mathbf{M}_i(x)
=
\Big[
\mathbf{m}^{(\ell)}_{i,s}(x)
\Big]_{\ell\in\mathcal{L},\, s=1}^{S}
\in\mathbb{R}^{|\mathcal{L}|\times S\times d_e}.
\label{eq:memory-tensor}
\end{equation}

In practice, these memory tensors can be precomputed offline for the training set and reused during injection-side learning, which decouples expert-side generation from frozen-base alignment. 
Compared with single-vector compression, this construction preserves richer specialist structure while retaining a constant-budget interface. 
Different slots capture different semantic regions of the generated background, while different layers preserve complementary abstraction levels.

\subsection{Latent Memory Injection Learning}
\label{sec:method-injection}

As shown in Figure~\ref{fig:gag-training}(d), the memory tensor in Eq.~\eqref{eq:memory-tensor} still lies in the expert representation space, and we therefore learn an injection-side module that transfers it into the embedding geometry of the frozen base model.

\textbf{Per-slot cross-layer fusion.}
For each slot $s$, GAG learns slot-specific layer-mixing weights. 
Let $\mathbf{w}_s\in\mathbb{R}^{|\mathcal{L}|}$ denote trainable logits. 
We compute
\begin{equation}
\beta_{s,\ell}
=
\frac{\exp(w_{s,\ell})}
{\sum_{\ell'\in\mathcal{L}}\exp(w_{s,\ell'})},
\label{eq:layermix-weight}
\end{equation}
and fuse the memory across layers as
\begin{equation}
\tilde{\mathbf{m}}_{i,s}(x)
=
\sum_{\ell\in\mathcal{L}}
\beta_{s,\ell}\mathbf{m}^{(\ell)}_{i,s}(x)
\in\mathbb{R}^{d_e}.
\label{eq:layermix}
\end{equation}
This per-slot formulation allows different memory slots to emphasize different representational depths.

\textbf{Gated residual projection.}
Each fused slot is then aligned to the frozen base-model embedding space through a gated residual projector. 
We first compute a base projection
\begin{equation}
\mathbf{u}_{i,s}(x)
=
\mathbf{W}_{\mathrm{in}}
\,\mathrm{LN}\!\left(\tilde{\mathbf{m}}_{i,s}(x)\right),
\label{eq:proj-base}
\end{equation}
followed by an update branch
\begin{equation}
\Delta\mathbf{u}_{i,s}(x)
=
\mathbf{W}_2\,
\mathrm{GELU}\!\left(
\mathbf{W}_1\,\mathrm{LN}\!\left(\mathbf{u}_{i,s}(x)\right)
\right),
\label{eq:proj-update}
\end{equation}
and a gating branch
\begin{equation}
\mathbf{g}_{i,s}(x)
=
\sigma\!\left(
\mathbf{W}_g\,\mathrm{LN}\!\left(\mathbf{u}_{i,s}(x)\right)
\right).
\label{eq:proj-gate}
\end{equation}
The final projected latent token is
\begin{equation}
\mathbf{z}_{i,s}(x)
=
\mathbf{u}_{i,s}(x)
+
\mathbf{g}_{i,s}(x)\odot\Delta\mathbf{u}_{i,s}(x)
\in\mathbb{R}^{d_b}.
\label{eq:proj-final}
\end{equation}
Compared with a plain MLP projector, this projector preserves coarse alignment through the residual path while allowing adaptive refinement through gated nonlinear updates.

\textbf{Constant-budget latent token injection.}
Let $s_{1:n}$ denote the answering prompt for $\mathrm{LLM}_{\mathrm{base}}$, containing $S$ reserved anchor positions $\{a_1,\ldots,a_S\}$. 
Let $\mathbf{E}_{\theta}(s_{1:n})\in\mathbb{R}^{n\times d_b}$ denote the input embeddings of the frozen base model. 
We replace the anchor embeddings with the projected latent slots:
\begin{equation}
\mathbf{E}^{(i)}_{\theta}(x)
=
\mathbf{E}_{\theta}(s_{1:n})
\quad
\text{with}
\quad
\mathbf{E}_{\theta}(s_{a_s})\leftarrow \mathbf{z}_{i,s}(x),
\;\; s=1,\ldots,S.
\label{eq:token-replace}
\end{equation}
The final answer is decoded by the frozen base model:
\begin{equation}
\hat{y}\sim p_{\theta}\!\left(y\mid \mathbf{E}^{(i)}_{\theta}(x)\right).
\label{eq:decode}
\end{equation}

\textbf{Learning objective.}
During this phase, we freeze both the base model $\theta$ and the domain expert $\phi_i$, and optimize only the injection-side parameters, including the layer-mixing weights, the projector, and an auxiliary semantic head.
The full injection-side training path is illustrated in Figure~\ref{fig:gag-training}(d).

The primary answer-generation objective is the negative log-likelihood
\begin{equation}
\mathcal{L}_{\mathrm{nll}}
=
-\sum_{t=1}^{|y|}
\log p_{\theta}\!\left(y_t\mid y_{<t},\mathbf{E}^{(i)}_{\theta}(x)\right).
\label{eq:loss-nll}
\end{equation}

To encourage semantic faithfulness beyond token-level matching, we introduce a latent semantic alignment loss. 
Let $\mathbf{H}^{\mathrm{ans}}_{\theta}(x)\in\mathbb{R}^{T_y\times d_b}$ denote the hidden states of the frozen base model over answer tokens, and let
\begin{equation}
\bar{\mathbf{h}}^{\mathrm{ans}}(x)
=
\mathrm{MeanPool}\!\left(\mathbf{H}^{\mathrm{ans}}_{\theta}(x)\right)
\label{eq:ans-pool}
\end{equation}
be the pooled answer representation. 
A semantic head $g(\cdot)$ maps it into a semantic space:
\begin{equation}
\mathbf{v}(x)=g\!\left(\bar{\mathbf{h}}^{\mathrm{ans}}(x)\right).
\label{eq:semantic-head}
\end{equation}
Let $e(y)$ be the representation of the gold answer obtained from a frozen semantic encoder. 
We define
\begin{equation}
\mathcal{L}_{\mathrm{sem}}
=
1-\cos\!\big(\mathbf{v}(x),e(y)\big).
\label{eq:loss-sem}
\end{equation}

Because multiple latent slots are injected simultaneously, we further regularize them to remain complementary rather than collapse into redundant replicas. 
For one training sample, the diversity loss is
\begin{equation}
\mathcal{L}_{\mathrm{div}}
=
\frac{1}{S(S-1)}
\sum_{s\neq s'}
\cos^2\!\big(\mathbf{z}_{i,s}(x),\mathbf{z}_{i,s'}(x)\big).
\label{eq:loss-div}
\end{equation}

The final objective is
\begin{equation}
\mathcal{L}
=
\alpha_{\mathrm{nll}}\mathcal{L}_{\mathrm{nll}}
+
\alpha_{\mathrm{sem}}\mathcal{L}_{\mathrm{sem}}
+
\alpha_{\mathrm{div}}\mathcal{L}_{\mathrm{div}},
\label{eq:loss-total}
\end{equation}
where $\alpha_{\mathrm{nll}}$, $\alpha_{\mathrm{sem}}$, and $\alpha_{\mathrm{div}}$ control the trade-off among generation fidelity, semantic alignment, and slot diversity.

\subsection{Prototype Plug-and-Play Routing}
\label{sec:method-ppr}

As illustrated in Figure~\ref{fig:gag-training}(e), to support unified mixed-domain deployment, GAG incorporates \textbf{Prototype Plug-and-Play Routing (PPR)}, a training-free router based on nearest-prototype matching in a frozen query-embedding space.

Let $g_{\eta}$ be a frozen encoder and $\mathrm{Pool}(\cdot)$ a fixed pooling operator. 
Each query is embedded and normalized as
\begin{equation}
\mathbf{e}(x)
=
\frac{\mathrm{Pool}(g_{\eta}(x))}
{\|\mathrm{Pool}(g_{\eta}(x))\|_2}.
\label{eq:ppr-embed}
\end{equation}
For each route $i\in\{0,1,\ldots,N\}$, including the general route and the private-domain routes, we cluster historical queries into a prototype bank
\begin{equation}
\mathbf{P}_i
=
\{\mathbf{p}_{i,1},\ldots,\mathbf{p}_{i,C_i}\}.
\label{eq:ppr-proto}
\end{equation}
At inference time, the routing score is computed by nearest-prototype similarity:
\begin{equation}
s_i(x)=\max_c\;\mathbf{e}(x)^\top \mathbf{p}_{i,c},
\qquad
r(x)=\arg\max_i s_i(x).
\label{eq:ppr-route}
\end{equation}
Figure~\ref{fig:gag-training}(e) visualizes both the offline prototype-bank construction and the online nearest-prototype routing procedure.
The selected route either invokes the general base model directly or activates the corresponding domain-specific module. 
A key advantage of PPR is modularity: adding a new domain only requires its expert module and prototype bank, while leaving the frozen base model and existing routes unchanged.

\begin{table*}[t]
\centering
\small
\setlength{\tabcolsep}{3.0pt}
\renewcommand{\arraystretch}{1.12}
\caption{\textbf{Overall performance in the unified mixed-domain setting.} 
We report results on the Materials domain, the Adjuvant domain, and the general domain. 
BertScore and StsScore are reported on a $\times 100$ scale. 
Numbers in parentheses for BertScore and StsScore denote relative improvements over Base-Model-Only in the corresponding domain. 
\textbf{Bold} indicates the best result and \underline{underline} indicates the second best.}
\label{tab:overall-performance}
\resizebox{\textwidth}{!}{%
\begin{tabular}{@{}lccccccccccccc@{}}
\toprule
\multirow{2}{*}{\textbf{System}} 
& \multicolumn{6}{c}{\textbf{Materials Domain}} 
& \multicolumn{6}{c}{\textbf{Adjuvant Domain}} 
& \multicolumn{1}{c}{\textbf{General Domain}} \\
\cmidrule(lr){2-7} \cmidrule(lr){8-13} \cmidrule(l){14-14}
& \textbf{BertScore} 
& \textbf{StsScore}
& \textbf{Reason.}
& \textbf{Acc.}
& \textbf{Usab.}
& \textbf{LLM Avg}
& \textbf{BertScore}
& \textbf{StsScore}
& \textbf{SS}
& \textbf{RS}
& \textbf{IS}
& \textbf{LLM Avg}
& \textbf{EM Avg} \\
\midrule
Base-Model-Only 
& 57.50 & 61.29 & 6.72 & 5.73 & 6.22 & 6.08 
& 53.58 & 74.97 & 5.17 & 6.38 & 5.28 & 5.61 
& \underline{42.16} \\

RAG 
& 61.14(6.33\%) & 82.31(34.30\%) & 7.42 & 6.40 & 6.89 & 6.75 
& 60.78(13.44\%) & \underline{81.84(9.16\%)} & 5.76 & 6.69 & 5.67 & 6.04 
& \textemdash \\

GraphRAG
& 62.35(8.43\%) & 82.74(35.00\%) & 7.68 & 6.56 & 7.02 & 6.92
& 61.11(14.05\%) & 79.33(5.82\%) & 5.85 & 6.98 & 5.52 & 6.12
& \textemdash \\

xRAG 
& 60.17(4.64\%) & 82.02(33.82\%) & 7.34 & 6.32 & 6.80 & 6.67 
& 58.03(8.31\%) & 78.11(4.19\%) & 5.70 & 6.71 & 5.21 & 5.87 
& \textemdash \\

Prompt-Tuning 
& \underline{67.10(16.70\%)} & 84.27(37.49\%) & 7.89 & 7.02 & 7.95 & 7.47 
& 62.40(16.46\%) & 79.25(5.71\%) & 5.83 & 7.07 & \underline{6.02} & 6.31 
& 35.64 \\

LoRA SFT 
& 67.08(16.66\%) & \underline{85.08(38.82\%)} & 7.77 & \underline{7.35} & \textbf{8.96} & \underline{7.92}
& \underline{63.26(18.07\%)} & 80.24(7.03\%) & \textbf{6.75} & 7.54 & 5.87 & \underline{6.72}
& 37.72 \\

BLADE 
& 66.55(15.74\%) & 84.35(37.62\%) & \underline{8.36} & 7.27 & 8.04 & 7.72 
& 62.71(17.04\%) & 79.84(6.50\%) & 6.15 & \underline{7.63} & 5.83 & 6.54 
& 33.96 \\

GAG (Ours) 
& \textbf{69.11(20.19\%)} & \textbf{87.26(42.37\%)} & \textbf{8.94} & \textbf{7.95} & \underline{8.80} & \textbf{8.40}
& \textbf{64.28(19.97\%)} & \textbf{82.37(9.87\%)} & \underline{6.52} & \textbf{8.07} & \textbf{6.97} & \textbf{7.19}
& \textbf{42.35} \\
\bottomrule
\end{tabular}%
}

\end{table*}

\section{Experimental Setup}
\label{sec:exp}

\subsection{Datasets and Metrics}
We evaluate GAG on both \textbf{general-domain QA} and \textbf{specialist private-domain QA} to quantify whether modular knowledge injection improves domain expertise without compromising broad usability. For general QA, we follow prior work and report performance on six widely used open-domain benchmarks—FreebaseQA \citep{jiang-etal-2019-freebaseqa}, HotpotQA \citep{yang2018hotpotqa}, Natural Questions \citep{kwiatkowski2019natural}, TriviaQA \citep{2017arXivtriviaqa}, WebQuestions \citep{berant-etal-2013-semantic}, and PopQA \citep{mallen2023llm_memorization}—using \textbf{Exact Match (EM)} with standard answer normalization, which provides a stringent measure of factual correctness under canonical string matching. To study domain knowledge injection, we focus on two specialist domains: \textbf{catalytic materials} and \textbf{immunology adjuvant}. Concretely, we treat \citep{chencatalystbench} and \citep{chen2026an} as the supervision sources for domain expert knowledge injection, and evaluate on their held-out test sets to quantify specialist QA quality. Because reference answers in these domains are often free-form and allow surface variation, we report \textbf{BERTScore} \citep{zhang2019bertscore} computed with SciBERT \citep{beltagy-etal-2019-scibert}, and \textbf{StsScore} \citep{reimers2019sentence} computed with \texttt{sentence-transformers/all-mpnet-base-v2}, to better reflect semantic faithfulness in technical language. In addition, we report benchmark-aligned \textbf{LLM-score} using \texttt{gpt-4o} as the judge model. 
For the materials domain, we follow the CatalystBench protocol and score answers along Reasonableness, Accuracy, and Usability, with the final score aggregated using the benchmark-defined weighting scheme (Reasonableness 20\%, Accuracy 50\%, Usability 30\%). For the adjuvant domain, we follow the benchmark protocol and score answers along Similarity Score (SS), Rationality Score (RS), and Inclusiveness Score (IS), with the final score computed, following the benchmark protocol, as their arithmetic mean. More detailed dataset statistics are provided in Appendix~\ref{app:data}.

\subsection{Implementation Details}
\label{sec:impl}
We instantiate the frozen base model $\mathrm{LLM}_{\text{base}}$ with \textbf{Qwen3-8B} and domain expert models $\mathrm{LLM}_{\text{domain},i}$ with \textbf{Qwen3-1.7B} \citep{yang2025qwen3}. 
Unless otherwise specified, GAG constructs latent memory from four expert layers, namely the last layer together with the second, fourth, and sixth layers below it, and compresses them into four memory slots for fixed-budget injection. We use \textbf{SciBERT} \citep{beltagy-etal-2019-scibert} as the frozen semantic encoder for semantic alignment. 
For routing, we use \textbf{Qwen3-1.7B} as a frozen query encoder and treat general as a peer route, requiring neither router training nor threshold tuning. 
All experiments are run on 8$\times$NVIDIA A100 GPUs with bfloat16 precision and FlashAttention-2~\citep{dao2023flashattention2}; full training hyperparameters, routing configuration, inference and decoding settings, and prompt templates are provided in Appendix~\ref{app:impl} and Appendix~\ref{app:prompts}.

\subsection{Baselines}
We compare against representative and competitive knowledge-injection baselines: \textbf{(i) Base-Model-Only}, where Qwen3-8B answers directly without any external knowledge; \textbf{(ii) RAG} \citep{lewis2020retrieval}, which builds domain corpora by parsing scientific papers with MinerU2.5 \citep{niu2025mineru25decoupledvisionlanguagemodel}, retrieves top-30 domain candidates via ColBERTv2 \citep{santhanam2022colbertv2}, reranks them with \texttt{bge-reranker-v2-m3}, and conditions the same base model on the top-$k$ retrieved passages; \textbf{(iii) GraphRAG} \citep{edge2024local}, which first uses GPT-4o to construct a graph index and associated community reports offline, then performs query-time local search over the graph-derived index, and finally uses Qwen3-8B to generate the answer; \textbf{(iv) xRAG} \citep{cheng2024xrag}, which performs retrieval augmentation by retrieving the top-1 background passage from the corresponding domain knowledge base and compressing it under an extreme budget; \textbf{(v) Prompt-Tuning} \citep{lester2021power}, which adapts the base model by prepending a small set of learnable soft prompt tokens, where we use 8 virtual tokens and a maximum sequence length of 2048; \textbf{(vi) LoRA SFT} \citep{hu2022lora}, which performs parameter-efficient supervised fine-tuning on the domain QA data using rank-8 LoRA adapters with $\alpha=16$ and a maximum sequence length of 2048, applied to the attention projection modules; and \textbf{(vii) BLADE} \citep{li2025blade}, which adopts a two-step explicit transfer pipeline where a smaller domain model initialized from Qwen3-1.7B and equipped with a learned soft prompt first generates domain knowledge, and the same base model then answers conditioned on the generated text.

\section{Experimental Results}
\label{sec:results}

\subsection{Overall Performance}
\label{sec:overall}

Table~\ref{tab:overall-performance} reports overall performance in the unified mixed-domain setting, covering two specialist scientific domains and six general-domain QA benchmarks under the same Qwen3-8B backbone. We omit RAG/GraphRAG/xRAG on the general-domain benchmarks because they are used here as closed-book regression checks: enabling open-domain retrieval would change the evaluation setting, while disabling retrieval would reduce them to the Base-Model-Only route. Overall, \textbf{GAG achieves the strongest specialist performance across both Materials and Adjuvant while preserving general-domain capability.}

The comparison with RAG, GraphRAG, and xRAG suggests that the main limitation in private scientific QA is not merely context budget, but the reliability of retrieved evidence itself. Although GraphRAG improves over standard RAG through graph-structured evidence organization, its gains remain limited, and xRAG stays close to or below standard RAG despite more aggressive compression. This indicates that chunk fragmentation, retrieval mismatch, and incomplete coverage remain the dominant bottlenecks. By avoiding retrieval-time text serialization and instead injecting specialist knowledge in latent form, GAG yields stronger and more stable specialist gains. Unless otherwise specified, the RAG results reported in Table~\ref{tab:overall-performance} are obtained with the best-performing retrieval depth, and the results under different top-$k$ settings are provided in Appendix~\ref{app:rag-topk}.

The comparison with Prompt-Tuning, LoRA SFT, and BLADE further highlights the practical advantage of GAG in mixed-domain deployment. Although these methods improve specialist performance over the base model to varying degrees, GAG still achieves the strongest results on both Materials and Adjuvant. Moreover, these baselines incur clear degradation on the general-domain benchmarks, whereas GAG preserves---and indeed slightly improves---the general-domain average over Base-Model-Only. Detailed results on the six individual general-domain benchmarks are provided in Appendix~\ref{app:general-domain-results}. This directly supports the central goal of GAG: enhancing specialist-domain capability without sacrificing the broad usability of the frozen base model.

\subsection{Routing Accuracy of PPR}
\label{sec:ppr-acc}

Reliable selective activation is a prerequisite for plug-and-play expert deployment, since misrouting can turn knowledge injection into harmful interference. Table~\ref{tab:ppr-routing-main} shows that \textbf{PPR achieves near-oracle routing under a fully frozen setup}: using a frozen Qwen3-1.7B encoder and nearest-prototype matching, PPR attains \textbf{99.72\%} micro-averaged accuracy for Gen+Mat, and remains \textbf{99.61\%} after incrementally adding Adj without modifying any existing routes. Per-route accuracy stays uniformly high, indicating that PPR provides a stable, non-parametric routing interface for scalable plug-and-play expert composition; additional routing results under broader incremental domain expansion are deferred to Appendix~\ref{app:ppr-routing}.

\begin{table}[t]
\centering
\small
\setlength{\tabcolsep}{4pt}
\renewcommand{\arraystretch}{1.12}
\caption{\textbf{Routing accuracy of PPR under incremental route expansion.}
Micro Acc. denotes micro-averaged routing accuracy over all evaluated queries. Per-route Acc. reports class-wise routing accuracy for each active route.}
\label{tab:ppr-routing-main}
\resizebox{\columnwidth}{!}{%
\begin{tabular}{@{}l l c c c c@{}}
\toprule
\textbf{Router Configuration} & \textbf{Active Routes} &
\textbf{Micro Acc. (\%)} &
\multicolumn{3}{c}{\textbf{Per-route Acc. (\%)}} \\
\cmidrule(lr){4-6}
& & & \textbf{Gen} & \textbf{Mat} & \textbf{Adj} \\
\midrule
PPR (2 routes) & Gen + Mat & 99.72 & 99.65 & 99.85 & \textemdash \\
PPR (3 routes) & Gen + Mat + Adj & 99.61 & 99.65 & 99.38 & 99.69 \\
\bottomrule
\end{tabular}%
}
\end{table}

\subsection{Efficiency Analysis}
\label{sec:efficiency}

Table~\ref{tab:efficiency-main} compares Base-Model-Only, RAG, and GAG on a fixed 128-query subset of the Adjuvant benchmark under the same Qwen3-8B decoding setup on NVIDIA A100 GPUs. Base-Model-Only answers directly with the base model under the same domain instruction and without external knowledge. RAG performs ColBERTv2 retrieval to obtain top-30 candidates, reranks the top-30 passages, and conditions the same base model on the top-5 passages, corresponding to the best-performing RAG setting on the Adjuvant benchmark. GAG uses the same Qwen3-8B backbone but injects four latent memory slots; its total cost includes both the small domain expert and the base model. We report approximate compute as TFLOPs/query based on prompt and generated token counts together with the model architecture.

The results show that GAG achieves the best trade-off between efficiency and effectiveness among the compared methods. It not only delivers the strongest specialist performance, but also uses the fewest added tokens and achieves the lowest total compute and latency. Compared with RAG, this advantage reflects the benefit of replacing long retrieved text and retrieval-side processing with a constant-budget latent interface. Compared with Base-Model-Only, GAG is also more efficient in our specialist-domain setting, which suggests that direct answering without effective specialist knowledge tends to produce less focused and longer generations for each query, leading to higher compute and latency. Taken together, these results further support the central claim of GAG: compact latent knowledge injection is not only more effective, but also more deployment-friendly than retrieval-time text conditioning for private scientific QA.

\begin{table}[H]
\centering
\small
\setlength{\tabcolsep}{3.8pt}
\renewcommand{\arraystretch}{1.10}
\caption{\textbf{Efficiency--effectiveness comparison.}
For GAG, the second line in each cost cell reports the decomposition of total cost into the small domain expert and the base model.}
\label{tab:efficiency-main}
\resizebox{\columnwidth}{!}{%
\begin{tabular}{@{}lccccc@{}}
\toprule
\textbf{Method} & \textbf{BertScore} & \textbf{StsScore} & \textbf{Added Tokens} & \textbf{TFLOPs/query} & \textbf{Latency (s/query)} \\
\midrule
Base-Model-Only & 53.58 & 74.97 & 0   & 11.8327 & 15.0998 \\
RAG             & 60.78 & 81.84 & 666.1 & 18.7270 & 11.8306 \\
GAG (Ours)      & \textbf{64.28} & \textbf{82.37} & 4 &
\begin{tabular}[c]{@{}c@{}}\textbf{8.7471}\\[-1pt]\scriptsize (1.6491 + 7.0980)\end{tabular} &
\begin{tabular}[c]{@{}c@{}}\textbf{10.6413}\\[-1pt]\scriptsize (4.0321 + 6.6092)\end{tabular} \\
\bottomrule
\end{tabular}%
}

\end{table}

\begin{figure*}[t]
\centering
\includegraphics[width=\textwidth]{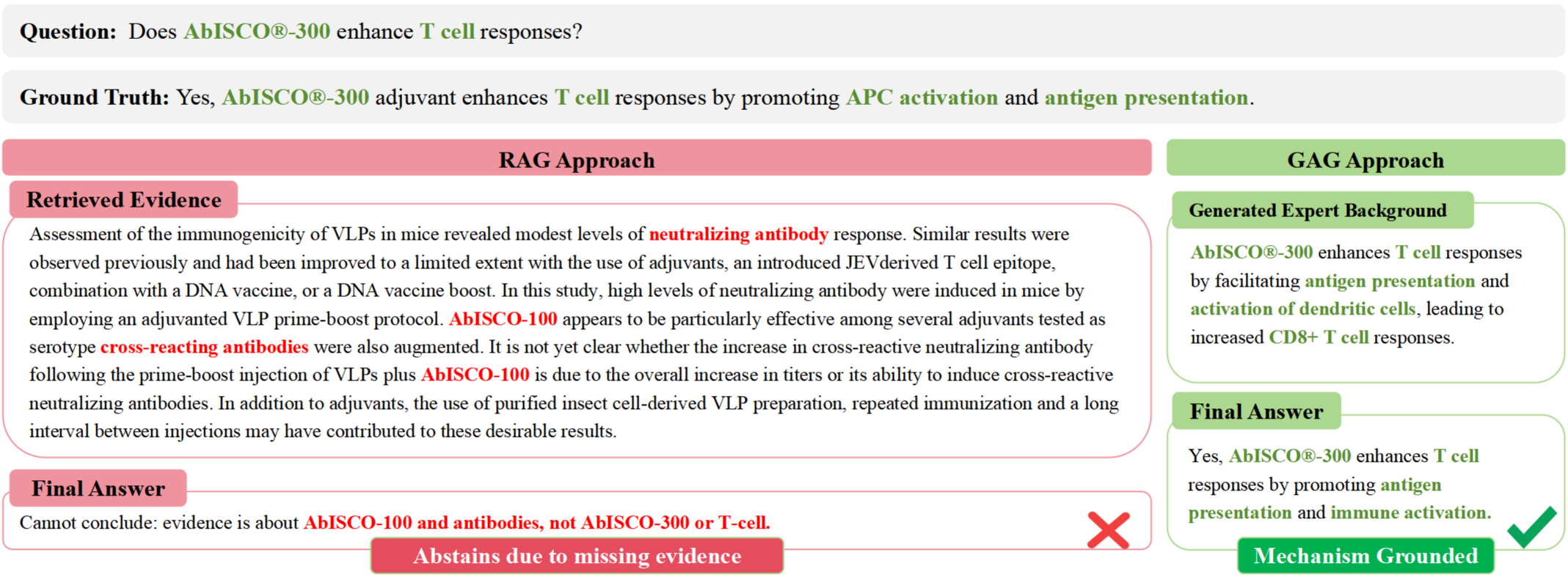}
\caption{\textbf{Case study (RAG vs.\ GAG).}
We contrast RAG’s retrieved evidence and answer with GAG’s latent-memory route and answer for the same query and reference.}
\label{fig:case-study}
\end{figure*}

\section{Analysis}
\label{sec:analysis}

\subsection{Ablation on Training Components}
\label{sec:ablation-training}

Table~\ref{tab:ablation-training-stages} examines the contribution of the three training components of GAG on the Materials domain: Domain-Adaptive Pretraining, Expert QA Specialization, and Latent Memory Injection Learning. Removing any component causes a clear performance drop, showing that all three are indispensable for effective specialist transfer. The two stages of domain expert construction play complementary roles: removing Domain-Adaptive Pretraining weakens the specialist prior of the expert model, while removing Expert QA Specialization reduces its ability to organize and express domain knowledge in a query-aware manner, making the resulting latent memory less informative. The largest degradation occurs when Latent Memory Injection Learning is removed, indicating that specialist knowledge extracted by the domain expert cannot be effectively utilized by the frozen base model without explicit cross-model alignment. Overall, the full model performs best, confirming that effective specialist transfer requires both strong domain expert construction and learned latent alignment.

\begin{table}[H]
\centering
\small
\setlength{\tabcolsep}{4.6pt}
\renewcommand{\arraystretch}{1.12}
\caption{\textbf{Ablation of the three training components of GAG.}}
\label{tab:ablation-training-stages}
\resizebox{\columnwidth}{!}{%
\begin{tabular}{@{}lcccc@{}}
\toprule
\textbf{Variant} & \textbf{Domain-Adaptive} & \textbf{Expert QA} & \textbf{Latent Memory Injection} & \textbf{BertScore} \\
& \textbf{Pretraining} & \textbf{Specialization} & \textbf{Learning} & \textbf{($\times 100$)} \\
\midrule
w/o Domain-Adaptive Pretraining     & $\times$     & $\checkmark$ & $\checkmark$ & 66.29 \\
w/o Expert QA Specialization        & $\checkmark$ & $\times$     & $\checkmark$ & 65.21 \\
w/o Latent Memory Injection Learning& $\checkmark$ & $\checkmark$ & $\times$     & 57.88 \\
Full GAG                            & $\checkmark$ & $\checkmark$ & $\checkmark$ & \textbf{69.11} \\
\bottomrule
\end{tabular}%
}
\end{table}

\subsection{Ablation on the Number of Memory Slots}
\label{sec:ablation-num-slots}

Table~\ref{tab:ablation-num-slots} studies the effect of the number of memory slots on the Materials domain. The results show that increasing the slot number from 1 to 4 consistently improves performance, indicating that a single compressed vector is insufficient to preserve the specialist knowledge needed for effective injection. At the same time, further increasing the number of slots from 4 to 8 brings no additional gain and instead leads to a slight drop. This suggests that while multi-slot memory is crucial for retaining complementary specialist signals, an excessively large slot budget may introduce redundancy and weaken the compactness of the latent interface. Overall, four memory slots provide the best trade-off between specialist knowledge preservation and efficient latent injection in GAG. 
Additional analyses on the learning objective design in latent memory injection learning, the layer-source configuration of background memory, expert-size scaling, base-model scaling, the full fine-tuning upper bound, and the cross-family transferability of GAG are provided in Appendix~\ref{app:additional-analysis}.

\begin{table}[H]
\centering
\small
\setlength{\tabcolsep}{7pt}
\renewcommand{\arraystretch}{1.12}
\caption{\textbf{Ablation on the number of memory slots.} 
$\Delta$ denotes the absolute BertScore difference from the full model with 4 memory slots. \textbf{Bold} indicates the best result and \underline{underline} indicates the second best.}
\label{tab:ablation-num-slots}
\resizebox{0.9\columnwidth}{!}{%
\begin{tabular}{@{}ccc@{}}
\toprule
\textbf{Number of Memory Slots} & \textbf{BertScore ($\times 100$)} & \textbf{$\Delta$ vs. Full} \\
\midrule
1 & 66.86 & -2.25 \\
2 & \underline{67.84} & -1.27 \\
4 & \textbf{69.11} & \textbf{0.00} \\
8 & 69.03 & -0.08 \\
\bottomrule
\end{tabular}%
}
\end{table}

\subsection{Case Study}
\label{sec:case-study}

Figure~\ref{fig:case-study} illustrates retrieval brittleness in chunked private corpora through an adjuvant-domain example: the query targets AbISCO-300 and a T-cell/APC mechanism, yet RAG’s evidence is \textbf{entity-mismatched} (AbISCO-100) and dominated by humoral readouts, so the frozen base model cannot ground the requested mechanism and effectively abstains. This example highlights that even topically related retrieved passages may fail to provide the specific specialist evidence needed for correct answering. GAG instead conditions the base model through a fixed-budget \textbf{multi-slot latent memory interface} distilled from the domain expert module, avoiding prompt-time evidence serialization and its associated coverage gaps. 
Importantly, the displayed “Generated Expert Background” in the figure is only an analysis-time probe for interpretability: in the actual GAG pipeline, $\mathrm{LLM}_{\text{domain},i}$ produces \textbf{no explicit text output} to the base model. Instead, the expert's multi-layer hidden states are compressed into latent memory slots and injected after cross-layer fusion and projector alignment. This case supports our claim that GAG mitigates retrieval fragmentation and entity mismatch while preserving a fixed and predictable specialist knowledge interface under a frozen base model. In Appendix~\ref{app:more-interesting-cases}, we include additional case studies and error analysis.

\section{Conclusion}
\label{sec:conclusion}

We proposed \textbf{GAG}, a retrieval-free and plug-and-play framework for injecting private, domain-specific knowledge into a frozen base model through a constant-budget latent interface distilled from lightweight domain expert models. By moving from text-level evidence serialization to representation-level knowledge transfer, GAG directly addresses the key limitations of prevailing paradigms: it mitigates RAG's brittleness under chunking, retrieval mismatch, and long-context pressure, while avoiding the iteration cost, deployment instability, and general-capability regression risks that often accompany fine-tuning in continuously evolving private domains. In a unified mixed-domain setting spanning two scientific private-domain QA benchmarks together with general-domain queries, GAG consistently improves specialist-domain performance while preserving general-domain capability, achieving highly reliable routing and a favorable efficiency--effectiveness trade-off. These results highlight a practical path toward modular, scalable, and governance-friendly private-knowledge deployment in real-world LLM systems.

\bibliographystyle{ACM-Reference-Format}
\bibliography{custom}

\clearpage
\appendix

\section{Dataset Details}
\label{app:data}

Table~\ref{tab:dataset-stats} summarizes the datasets used in our study.
For general-domain QA, we report EM on six public benchmarks—FreebaseQA \citep{jiang-etal-2019-freebaseqa}, HotpotQA \citep{yang2018hotpotqa}, Natural Questions \citep{kwiatkowski2019natural}, TriviaQA \citep{2017arXivtriviaqa}, WebQuestions \citep{berant-etal-2013-semantic}, and PopQA \citep{mallen2023llm_memorization}.
Because the official dev/test splits of these benchmarks are substantially larger than needed to characterize general QA behavior under a fixed inference setup, we evaluate on a single fixed random subset from each dataset’s official dev/test split and reuse the same subsets across all methods, ensuring a consistent, reproducible, and benchmark-balanced probe of general capability.

For specialist private-domain QA, we use two specialist domains, catalytic materials and immunology adjuvant, together with their corresponding supervision sources for domain expert knowledge injection \citep{chencatalystbench,chen2026an}, paired with private literature corpora (986 papers for materials domain; 813 papers for adjuvant domain).
We evaluate specialist QA with three complementary metrics: BERTScore \citep{zhang2019bertscore} computed using \textsc{SciBERT} \citep{beltagy-etal-2019-scibert}, StsScore \citep{reimers2019sentence} computed using \nolinkurl{sentence-transformers/all-mpnet-base-v2}, and benchmark-aligned LLM-score using \texttt{gpt-4o} as the judge model.
For the materials domain, we follow the CatalystBench protocol and report Reasonableness, Accuracy, and Usability, with the final LLM-score aggregated using the benchmark-defined weighting scheme (20\%, 50\%, 30\%).
For the adjuvant domain, we follow the benchmark protocol and report Similarity Score (SS), Rationality Score (RS), and Inclusiveness Score (IS), with the final LLM-score computed as their arithmetic mean.

\begin{table*}[t]
\centering
\footnotesize
\setlength{\tabcolsep}{4pt}
\renewcommand{\arraystretch}{1.12}

\caption{\textbf{Dataset summary.} ``Eval'' denotes the number of evaluated questions used for each general-domain QA benchmark. Specialist private-domain QA benchmarks are paired with private literature corpora and evaluated using BERTScore, StsScore, and benchmark-aligned LLM-score.}
\label{tab:dataset-stats}

\begin{tabularx}{\textwidth}{@{}
  >{\raggedright\arraybackslash}p{0.17\textwidth}
  Y
  Y
  >{\centering\arraybackslash}p{0.22\textwidth}
@{}}
\toprule
\textbf{Category} & \textbf{Dataset} & \textbf{Split / Size} & \textbf{Metric} \\
\midrule
\multirow{6}{=}[-1ex]{\makecell[l]{General-domain QA}}
& FreebaseQA \citep{jiang-etal-2019-freebaseqa} & Eval: 1135 & EM \\
& HotpotQA \citep{yang2018hotpotqa} & Eval: 1135 & EM \\
& Natural Questions \citep{kwiatkowski2019natural} & Eval: 1135 & EM \\
& TriviaQA \citep{2017arXivtriviaqa} & Eval: 1135 & EM \\
& WebQuestions \citep{berant-etal-2013-semantic} & Eval: 1135 & EM \\
& PopQA \citep{mallen2023llm_memorization} & Eval: 1135 & EM \\
\midrule
\multirow{2}{=}[-2.5ex]{\makecell[l]{Specialist private-\\domain QA}}
& Catalytic Materials \citep{chencatalystbench}
& \makecell[l]{Train: 3{,}661\\ Test: 646\\ Corpus: 986 papers}
& \multirow{2}{=}[-2ex]{\makecell[c]{BERTScore (\textsc{SciBERT})\\StsScore (all-mpnet-base-v2)\\LLM-score (\texttt{gpt-4o})}} \\
& Immunology Adjuvant \citep{chen2026an}
& \makecell[l]{Train: 21{,}614\\ Test: 1{,}294\\ Corpus: 813 papers}
& \\
\bottomrule
\end{tabularx}
\end{table*}

\section{Additional Implementation Details}
\label{app:impl}

All experiments are run on 8$\times$NVIDIA A100 GPUs with bfloat16 precision, with FlashAttention-2 enabled when available. 
For reproducibility, we report the key hyperparameter settings for \textbf{Domain-Adaptive Pretraining}, \textbf{Expert QA Specialization}, and \textbf{Latent Memory Injection Learning} in Tables~\ref{tab:dapt-hparams}, \ref{tab:qa-hparams}, and \ref{tab:injection-hparams}, respectively.

For PPR, query representations are obtained by attention-masked mean pooling over the encoder's last-layer states followed by $\ell_2$ normalization. Prototype banks are built using 32 prototypes per domain, subsampling up to 10k in-domain queries for clustering.

Unless otherwise stated, we disable the model's explicit ``thinking'' mode during both training and evaluation (i.e., we use the non-thinking chat format). At inference, we apply nucleus/top-$k$ sampling for both background synthesis and answer generation with temperature $0.7$, top-$p$ $0.8$, and top-$k$ 20; the maximum generation lengths follow the benchmark configuration.

\begin{table}[t]
\centering
\small
\setlength{\tabcolsep}{6pt}
\renewcommand{\arraystretch}{1.08}
\caption{\textbf{Key hyperparameters for Domain-Adaptive Pretraining (Materials domain).}}
\label{tab:dapt-hparams}
\begin{tabular}{@{}l l@{}}
\toprule
\textbf{Hyperparameter} & \textbf{Value} \\
\midrule
Block size & 1024 \\
Per-device train batch size & 1 \\
Gradient accumulation steps & 8 \\
Learning rate & $5\times 10^{-5}$ \\
Weight decay & 0.01 \\
Training epochs & 2 \\
Warmup ratio & 0.03 \\
Precision & bfloat16 \\
Validation split ratio & 0.03 \\
Seed & 980406 \\
\bottomrule
\end{tabular}
\end{table}

\begin{table}[t]
\centering
\small
\setlength{\tabcolsep}{6pt}
\renewcommand{\arraystretch}{1.08}
\caption{\textbf{Key hyperparameters for Expert QA Specialization (Materials domain).}}
\label{tab:qa-hparams}
\begin{tabular}{@{}l l@{}}
\toprule
\textbf{Hyperparameter} & \textbf{Value} \\
\midrule
Max sequence length & 2048 \\
Per-device train batch size & 1 \\
Gradient accumulation steps & 8 \\
Learning rate & $5\times 10^{-6}$ \\
Weight decay & 0.0 \\
Training epochs & 3 \\
Warmup ratio & 0.03 \\
Precision & bfloat16 \\
Seed & 980406 \\
\bottomrule
\end{tabular}
\end{table}

\begin{table}[t]
\centering
\small
\setlength{\tabcolsep}{6pt}
\renewcommand{\arraystretch}{1.08}
\caption{\textbf{Key hyperparameters for Latent Memory Injection Learning (Materials domain).}}
\label{tab:injection-hparams}
\begin{tabular}{@{}l l@{}}
\toprule
\textbf{Hyperparameter} & \textbf{Value} \\
\midrule
Max sequence length & 2048 \\
Per-device train batch size & 1 \\
Gradient accumulation steps & 4 \\
Learning rate & $1\times 10^{-4}$ \\
Weight decay & 0.0 \\
LR scheduler & linear \\
Warmup ratio & 0.1 \\
Clip grad norm & 1.0 \\
Layer keys & \{last, $-2$, $-4$, $-6$\} \\
Number of memory slots & 4 \\
Slot pooling & segment\_softmax \\
Slot pooling temperature & 1.0 \\
Memory slot dropout & 0.0 \\
$\alpha_{\mathrm{nll}}, \alpha_{\mathrm{sem}}, \alpha_{\mathrm{div}}$ & $1.0,\ 0.01,\ 0.001$ \\
Seed & 980406 \\
\bottomrule
\end{tabular}
\end{table}

\section{Prompt Templates}
\label{app:prompts}

\begin{figure*}[t]
  \centering
  \includegraphics[width=\textwidth]{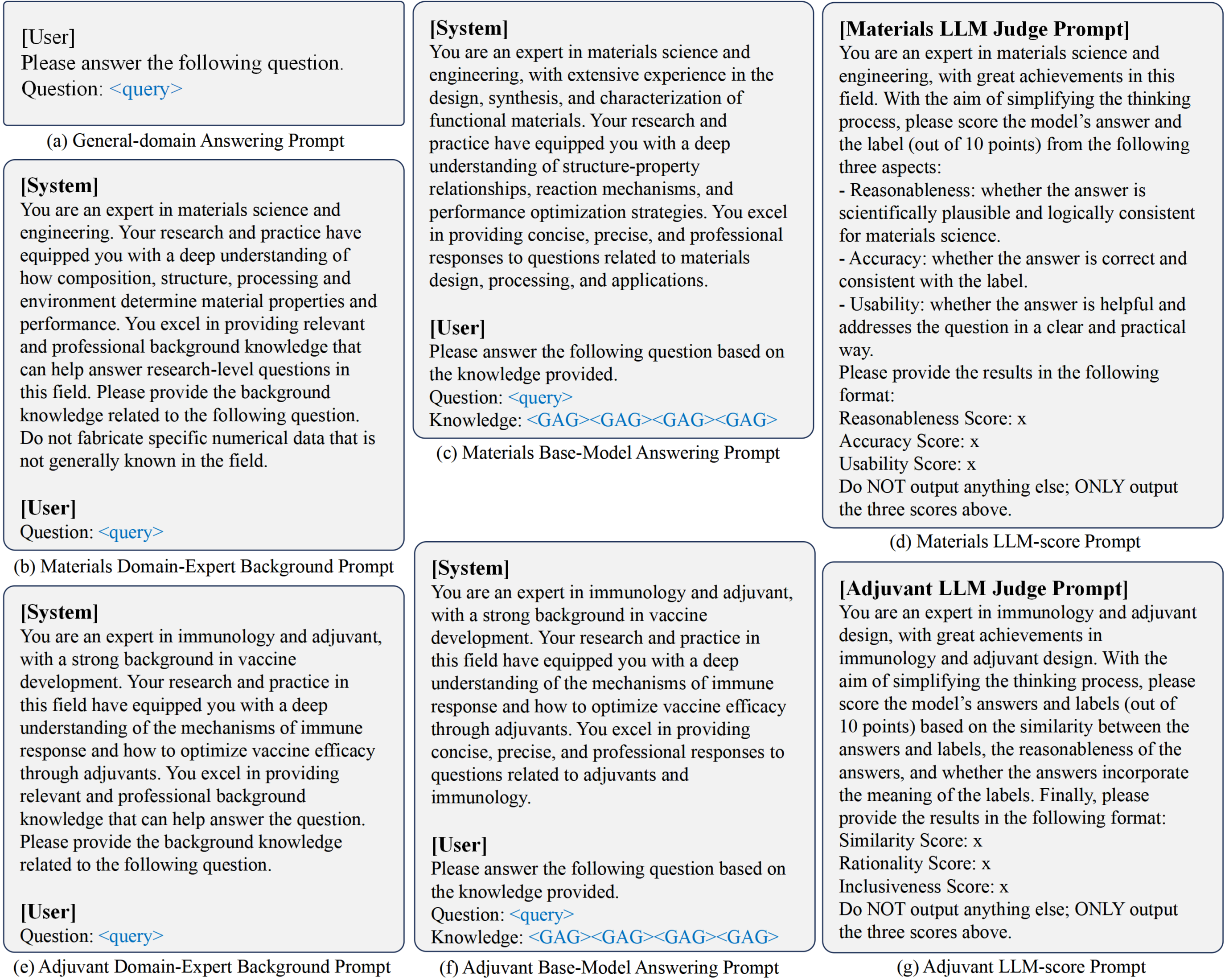}
  \caption{\textbf{Prompt templates used in GAG and benchmark-aligned evaluation.}
  (a) General-domain Answering Prompt.
  (b) Materials Domain-Expert Background Prompt.
  (c) Materials Base-Model Answering Prompt.
  (d) Materials LLM-score Prompt.
  (e) Adjuvant Domain-Expert Background Prompt.
  (f) Adjuvant Base-Model Answering Prompt.
  (g) Adjuvant LLM-score Prompt.
  Blue text denotes runtime placeholders.}
  \label{fig:prompt-templates}
\end{figure*}

Figure~\ref{fig:prompt-templates} summarizes the prompt templates used in our experiments.
To keep evaluation stable and reproducible, we use a small set of fixed chat-style templates and fill only the runtime placeholders (shown in blue in the figure).
Unless otherwise specified, all prompts use the non-thinking chat format.

For the \textbf{general route}, the frozen base model answers the query directly using a minimal answering template.
For each \textbf{specialist route}, prompting is organized into two stages: a \textbf{domain-expert background prompt} used to elicit question-conditioned specialist background, and a \textbf{base-model answering prompt} used by the frozen base model to produce the final answer under latent-memory injection.
In the answering prompts, the repeated \texttt{<GAG>} placeholders indicate the reserved anchor positions for injected latent memories.

We also include the \textbf{LLM-score prompts} used for benchmark-aligned evaluation in the materials and adjuvant domains.
These judge prompts follow the domain-specific evaluation criteria described in the main paper and provide the score dimensions used in our reported LLM-score results.

\section{RAG Sensitivity to Retrieval Depth}
\label{app:rag-topk}

Table~\ref{tab:rag_topk_vs_gag} further examines how retrieval depth affects RAG in the two specialist domains by varying $k$, the number of retrieved passages concatenated to the base-model prompt.
For brevity, Table~\ref{tab:overall-performance} reports the best-performing retrieval depth for each domain, while we provide the full sweep here.

\textbf{Discussion.}
RAG does not improve monotonically as $k$ increases.
Instead, performance first improves and then declines: the best results are obtained at $k{=}3$ on the Materials domain and at $k{=}5$ on the Adjuvant domain.
This trend suggests a trade-off between evidence coverage and context interference.
When $k$ is too small, relevant evidence may be missed; when $k$ becomes too large, additional retrieved passages introduce more distractors and increase competition within the prompt, which is particularly problematic in chunked private scientific corpora.
By contrast, GAG avoids prompt-time growth in retrieved text and remains consistently stronger under a fixed latent knowledge budget.

\begin{table}[t]
\centering
\small
\setlength{\tabcolsep}{4.2pt}
\renewcommand{\arraystretch}{1.10}
\caption{\textbf{Sensitivity of RAG to retrieval depth.}
\textbf{Bold} indicates the best overall result and \underline{underline} indicates the best RAG result within each metric.}
\label{tab:rag_topk_vs_gag}
\resizebox{\columnwidth}{!}{%
\begin{tabular}{@{}l c cc cc@{}}
\toprule
\multirow{2}{*}{\textbf{System}} 
& \multirow{2}{*}{\textbf{Top-$k$}} 
& \multicolumn{2}{c}{\textbf{Materials Domain}} 
& \multicolumn{2}{c}{\textbf{Adjuvant Domain}} \\
\cmidrule(lr){3-4} \cmidrule(l){5-6}
& 
& \textbf{BertScore} 
& \textbf{StsScore} 
& \textbf{BertScore} 
& \textbf{StsScore} \\
\midrule
Base-Model-Only 
& 0 
& 57.50 & 61.29 
& 53.58 & 74.97 \\
\midrule
\multirow{5}{*}{RAG}
& 1 
& 60.22 & 81.61 
& 60.01 & 79.30 \\
& 3 
& \underline{61.14} & \underline{82.31} 
& 60.63 & 81.07 \\
& 5 
& 60.38 & 81.98 
& \underline{60.78} & \underline{81.84} \\
& 7 
& 60.31 & 81.88 
& 60.48 & 81.65 \\
& 9 
& 60.70 & 82.17 
& 60.59 & 81.57 \\
\midrule
GAG (Ours) 
& -- 
& \textbf{69.11} & \textbf{87.26} 
& \textbf{64.28} & \textbf{82.37} \\
\bottomrule
\end{tabular}%
}
\end{table}

\section{Additional General-domain Benchmark Results}
\label{app:general-domain-results}

Table~\ref{tab:general-domain-results} reports the full per-benchmark EM results on the six general-domain QA datasets corresponding to the average reported in Table~\ref{tab:overall-performance}. 
Consistent with the main results, the specialist adaptation baselines substantially degrade general-domain capability, whereas GAG preserves it and attains the highest overall average.

\begin{table*}[t]
\centering
\small
\setlength{\tabcolsep}{4.2pt}
\renewcommand{\arraystretch}{1.10}
\caption{\textbf{Per-benchmark results on the six general-domain QA datasets.}
All numbers are Exact Match (EM). 
This table expands the general-domain average reported in Table~\ref{tab:overall-performance}. 
\textbf{Bold} indicates the best result and \underline{underline} indicates the second best.}
\label{tab:general-domain-results}
\resizebox{\textwidth}{!}{%
\begin{tabular}{@{}lccccccc@{}}
\toprule
\textbf{System} 
& \textbf{FreebaseQA} 
& \textbf{HotpotQA} 
& \textbf{Natural Questions} 
& \textbf{TriviaQA} 
& \textbf{WebQuestions} 
& \textbf{PopQA} 
& \textbf{Average} \\
\midrule
Base-Model-Only
& 61.06
& \underline{28.72}
& \textbf{35.15}
& \textbf{54.36}
& \underline{49.96}
& \underline{23.70}
& \underline{42.16} \\

Prompt-Tuning
& 60.70
& 21.32
& 25.99
& 51.28
& 37.27
& 17.27
& 35.64 \\

LoRA SFT
& \textbf{62.03}
& 24.41
& 28.46
& 52.51
& 39.47
& 19.47
& 37.72 \\

BLADE
& 59.56
& 20.18
& 25.81
& 42.20
& 36.21
& 19.82
& 33.96 \\

GAG (Ours)
& \underline{61.41}
& \textbf{29.16}
& \underline{34.98}
& \underline{53.57}
& \textbf{50.31}
& \textbf{24.67}
& \textbf{42.35} \\
\bottomrule
\end{tabular}%
}
\end{table*}

\section{Additional Results: Incremental Multi-domain Routing with PPR}
\label{app:ppr-routing}

Section~\ref{sec:ppr-acc} shows that PPR enables reliable selective activation in the main routed setting.
Here we further examine incremental multi-domain expansion under a strictly plug-and-play protocol.
Starting from General+Materials, we progressively add Adjuvant, Aviation \citep{agarwal2022there}, Law \citep{zheng2025reasoning}, and Math \citep{cobbe2021gsm8k}.
Crucially, at each step we construct and load only the prototype bank of the newly introduced route, while keeping the query encoder and all previously deployed prototype banks fully frozen.
Routing is performed by nearest-prototype cosine similarity, so performance directly reflects how well domain query manifolds remain separable in a shared embedding space without router retraining or threshold tuning.

\begin{table}[t]
\centering
\small
\setlength{\tabcolsep}{5pt}
\renewcommand{\arraystretch}{1.10}
\caption{\textbf{Evaluation pool for incremental multi-domain routing.}
The General route is a balanced union of six public QA subsets (sizes in parentheses).}
\label{tab:ppr-routing-data}
\begin{tabular}{@{}l l r@{}}
\toprule
\textbf{Route} & \textbf{Source / composition} & \textbf{\#Queries} \\
\midrule
General &
\makecell[l]{FreebaseQA (189) \\ HotpotQA (196) \\ Natural Questions (193) \\ TriviaQA (184) \\ WebQuestions (183) \\ PopQA (190)} &
1{,}135 \\
Materials & Materials specialist QA & 646 \\
Adjuvant & Adjuvant specialist QA & 1{,}294 \\
Aviation & sakharamg/AviationQA & 1{,}135 \\
Law & reglabs/housing\_qa & 1{,}135 \\
Math & openai/gsm8k & 1{,}135 \\
\midrule
\textbf{Total} & \textemdash & \textbf{6{,}480} \\
\bottomrule
\end{tabular}
\end{table}

Table~\ref{tab:ppr-routing-data} summarizes the evaluation pool spanning six routes, while Table~\ref{tab:ppr-routing-appendix} reports routing accuracy as the active route set grows from 2 to 6.
PPR exhibits strong scalability under incremental expansion: micro-averaged accuracy remains above $99.5\%$ across all stages, despite repeatedly increasing the decision space.
Moreover, per-route accuracy stays uniformly high for both newly added routes and previously deployed routes, with no meaningful degradation as new prototype banks are attached.
Together, these results position PPR as a non-parametric, deployment-friendly routing interface for modular specialist systems: domain expansion is realized by a lightweight prototype update, while the frozen base model and existing routes remain unchanged.

\begin{table}[t]
\centering
\small
\setlength{\tabcolsep}{4pt}
\renewcommand{\arraystretch}{1.08}
\caption{\textbf{Incremental multi-domain scalability of PPR.}
We progressively add new routes by loading only the corresponding prototype bank (the encoder and all existing routes remain frozen).
Micro acc. is overall routing accuracy; Per-route acc. is class-wise accuracy for each active route.}
\label{tab:ppr-routing-appendix}
\resizebox{\columnwidth}{!}{%
\begin{tabular}{@{}c l c c c c c c c@{}}
\toprule
\textbf{\#Routes} & \textbf{New domain} &
\textbf{Micro} &
\multicolumn{6}{c}{\textbf{Per-route acc. (\%)}} \\
\cmidrule(lr){4-9}
& & \textbf{acc. (\%)} &
\textbf{Gen} & \textbf{Mat} & \textbf{Adj} & \textbf{Avi} & \textbf{Law} & \textbf{Math} \\
\midrule
2 & \textemdash & 99.72 & 99.65 & 99.85 & \textemdash & \textemdash & \textemdash & \textemdash \\
3 & + Adj  & 99.61 & 99.65 & 99.38 & 99.69 & \textemdash & \textemdash & \textemdash \\
4 & + Avi  & 99.67 & 99.47 & 99.38 & 99.69 & 100.00 & \textemdash & \textemdash \\
5 & + Law  & 99.74 & 99.47 & 99.38 & 99.69 & 100.00 & 100.00 & \textemdash \\
6 & + Math & 99.74 & 99.47 & 99.38 & 99.69 & 100.00 & 100.00 & 99.74 \\
\bottomrule
\end{tabular}%
}
\end{table}

\section{Additional Analysis}
\label{app:additional-analysis}

\subsection{Ablation on the Learning Objective in Latent Memory Injection Learning}
\label{app:analysis-objective}

\begin{table}[t]
\centering
\small
\setlength{\tabcolsep}{5pt}
\renewcommand{\arraystretch}{1.10}
\caption{\textbf{Ablation on the learning objective in Latent Memory Injection Learning (Materials domain).}
$\Delta$ denotes the absolute BertScore difference from the full objective.
\textbf{Bold} indicates the best result and \underline{underline} indicates the second best.}
\label{tab:objective-ablation}
\resizebox{\columnwidth}{!}{%
\begin{tabular}{@{}lcccccc@{}}
\toprule
\textbf{Variant} 
& $\mathcal{L}_{\mathrm{nll}}$ 
& $\mathcal{L}_{\mathrm{sem}}$ 
& $\mathcal{L}_{\mathrm{div}}$ 
& \textbf{BertScore} 
& \textbf{$\Delta$ vs. Full} \\
&  &  &  & \textbf{($\times 100$)} & \\
\midrule
Only NLL                  & $\checkmark$ & $\times$     & $\times$     & 68.43 & -0.68 \\
NLL + Semantic            & $\checkmark$ & $\checkmark$ & $\times$     & \underline{68.87} & -0.24 \\
Full Objective            & $\checkmark$ & $\checkmark$ & $\checkmark$ & \textbf{69.11} & \textbf{0.00} \\
\bottomrule
\end{tabular}%
}
\end{table}

Table~\ref{tab:objective-ablation} studies the contribution of the three loss terms used in \textbf{Latent Memory Injection Learning}.
Starting from the NLL-only setting, adding the semantic alignment loss improves BertScore from 68.43 to 68.87, and further adding the diversity regularizer brings the score to 69.11.
These results show that the full three-term objective is the most effective: \(\mathcal{L}_{\mathrm{nll}}\) provides the core answer-generation signal, \(\mathcal{L}_{\mathrm{sem}}\) improves answer-level semantic faithfulness, and \(\mathcal{L}_{\mathrm{div}}\) further regularizes the injected latent slots to remain complementary rather than collapse into redundant replicas.

\subsection{Ablation on the Layer-source Configuration of Latent Memory Construction}
\label{app:analysis-layers}

Table~\ref{tab:layer-source-ablation} compares the default multi-layer memory construction against a last-layer-only variant.
Using only the last expert layer yields a BertScore of 68.58, whereas the full multi-layer setting with LayerMix reaches 69.11.
This result shows that collecting latent memory from multiple expert layers and performing cross-layer fusion is beneficial: the intermediate layers provide complementary specialist signals that are not fully preserved by the last layer alone.

\begin{table}[H]
\centering
\small
\setlength{\tabcolsep}{5pt}
\renewcommand{\arraystretch}{1.10}
\caption{\textbf{Ablation on the layer-source configuration of latent memory construction (Materials domain).}
$\Delta$ denotes the absolute BertScore difference from the full multi-layer setting.
\textbf{Bold} indicates the best result.}
\label{tab:layer-source-ablation}
\resizebox{\columnwidth}{!}{%
\begin{tabular}{@{}lccccl@{}}
\toprule
\textbf{Variant} 
& \textbf{Expert Layers} 
& \textbf{LayerMix} 
& \textbf{BertScore} 
& \textbf{$\Delta$ vs. Full} \\
&  &  & \textbf{($\times 100$)} & \\
\midrule
Last-layer only             & \{last\}                & $\times$     & 68.58 & -0.53 \\
Multi-layer + LayerMix (Full) & \{last, $-2$, $-4$, $-6$\} & $\checkmark$ & \textbf{69.11} & \textbf{0.00} \\
\bottomrule
\end{tabular}%
}
\end{table}

\subsection{Ablation on Domain-Expert Scaling}
\label{app:expert-scaling}

\begin{table}[b]
\centering
\small
\setlength{\tabcolsep}{5pt}
\renewcommand{\arraystretch}{1.10}
\caption{\textbf{Ablation on domain-expert scaling in the Materials domain.}
The frozen base model is fixed as Qwen3-8B.
$\Delta$ denotes the absolute improvement over the smallest expert (Qwen3-0.6B).
All scores are reported on a $\times 100$ scale.
\textbf{Bold} indicates the best result and \underline{underline} indicates the second best.}
\label{tab:expert-scaling}
\resizebox{0.95\columnwidth}{!}{%
\begin{tabular}{@{}lccccc@{}}
\toprule
\multirow{2}{*}{\textbf{Domain Expert}} & \multirow{2}{*}{\textbf{Frozen Base}} &
\multicolumn{2}{c}{\textbf{BertScore}} &
\multicolumn{2}{c}{\textbf{StsScore}} \\
\cmidrule(lr){3-4}\cmidrule(l){5-6}
& & \textbf{Score} & \textbf{$\Delta$} & \textbf{Score} & \textbf{$\Delta$} \\
\midrule
Qwen3-0.6B & Qwen3-8B & 68.98 & 0.00 & 87.09 & 0.00 \\
Qwen3-1.7B & Qwen3-8B & \underline{69.11} & +0.13 & \underline{87.26} & +0.17 \\
Qwen3-4B   & Qwen3-8B & \textbf{69.37} & \textbf{+0.39} & \textbf{87.95} & \textbf{+0.86} \\
\bottomrule
\end{tabular}%
}
\end{table}

Table~\ref{tab:expert-scaling} studies the effect of scaling the domain expert while keeping the frozen base model fixed as Qwen3-8B. We observe a consistent upward trend as the expert increases from Qwen3-0.6B to Qwen3-4B, indicating that a stronger domain expert can provide higher-quality question-conditioned specialist signals for latent memory construction. This is consistent with the role of the expert in GAG: a larger expert is better able to organize domain knowledge into more informative hidden trajectories before compression and projection. At the same time, the improvement remains moderate rather than dramatic, suggesting that the final performance is not determined by expert capacity alone. Under the constant-budget latent interface of GAG, the benefit of scaling the domain expert is ultimately bounded by the fixed capacity of the frozen base model to absorb and utilize the injected specialist memory, leading to diminishing returns once the expert-side signals become sufficiently informative.

\subsection{Ablation on Base-Model Scaling}
\label{app:base-scaling}

Table~\ref{tab:base-scaling} studies the effect of scaling the frozen base model while fixing the domain expert as Qwen3-1.7B. The results improve steadily from Qwen3-8B to Qwen3-32B, showing that larger base models can make better use of the injected latent memories and more effectively combine them with their stronger pretrained prior. This trend supports the decoupled design of GAG: once specialist knowledge has been distilled into latent memories, a stronger frozen backbone can exploit the same injected signals more effectively without changing the expert-side pipeline. Meanwhile, the gain is still limited in magnitude, which suggests that performance is also bounded by the quality and capacity of the fixed expert-side memory source. In other words, once the expert is fixed, enlarging the base model mainly improves downstream utilization of the injected knowledge, but cannot fully compensate for the information bottleneck on the expert side.

\begin{table}[H]
\centering
\small
\setlength{\tabcolsep}{5pt}
\renewcommand{\arraystretch}{1.10}
\caption{\textbf{Ablation on base-model scaling in the Materials domain.}
The domain expert is fixed as Qwen3-1.7B.
$\Delta$ denotes the absolute improvement over the default/smallest base model (Qwen3-8B).
All scores are reported on a $\times 100$ scale.
\textbf{Bold} indicates the best result and \underline{underline} indicates the second best.}
\label{tab:base-scaling}
\resizebox{0.95\columnwidth}{!}{%
\begin{tabular}{@{}lccccc@{}}
\toprule
\multirow{2}{*}{\textbf{Domain Expert}} & \multirow{2}{*}{\textbf{Frozen Base}} &
\multicolumn{2}{c}{\textbf{BertScore}} &
\multicolumn{2}{c}{\textbf{StsScore}} \\
\cmidrule(lr){3-4}\cmidrule(l){5-6}
& & \textbf{Score} & \textbf{$\Delta$} & \textbf{Score} & \textbf{$\Delta$} \\
\midrule
Qwen3-1.7B & Qwen3-8B  & 69.11 & 0.00 & 87.26 & 0.00 \\
Qwen3-1.7B & Qwen3-14B & \underline{69.19} & +0.08 & \underline{87.43} & +0.17 \\
Qwen3-1.7B & Qwen3-32B & \textbf{69.33} & \textbf{+0.22} & \textbf{87.78} & \textbf{+0.52} \\
\bottomrule
\end{tabular}%
}
\end{table}

\subsection{Upper-Bound Comparison with Full Fine-Tuning}
\label{app:upper-bound}

To contextualize the specialist gains of GAG, we further compare it against a full fine-tuning upper bound using the same Qwen3-8B backbone. Table~\ref{tab:upper-bound} shows that GAG approaches the fully fine-tuned Qwen3-8B upper bound remarkably closely on both specialist domains. These results suggest that a large portion of the attainable specialist gain can be recovered through latent memory injection alone, without sacrificing the frozen-base constraint, modular deployment, or general-domain capability.

\begin{table}[H]
\centering
\small
\setlength{\tabcolsep}{3.6pt}
\renewcommand{\arraystretch}{1.08}
\caption{\textbf{Upper-bound comparison with full fine-tuning on Qwen3-8B.} 
All scores are reported on a $\times 100$ scale.}
\label{tab:upper-bound}
\resizebox{\columnwidth}{!}{%
\begin{tabular}{@{}lcccc@{}}
\toprule
\multirow{2}{*}{\textbf{System}} 
& \multicolumn{2}{c}{\textbf{Materials}} 
& \multicolumn{2}{c}{\textbf{Adjuvant}} \\
\cmidrule(lr){2-3} \cmidrule(l){4-5}
& \textbf{BertScore} & \textbf{StsScore}
& \textbf{BertScore} & \textbf{StsScore} \\
\midrule
Base-Model-Only 
& 57.50 & 61.29 & 53.58 & 74.97 \\

GAG (Ours) 
& \underline{69.11} & \underline{87.26} & \underline{64.28} & \underline{82.37} \\

Full FT (Qwen3-8B, UB) 
& \textbf{70.15} & \textbf{88.09} & \textbf{65.05} & \textbf{83.22} \\
\bottomrule
\end{tabular}%
}
\end{table}

\subsection{Cross-family Transferability}
\label{app:analysis-cross-family}

Table~\ref{tab:cross-family-transfer} examines whether GAG remains effective when the domain expert and the frozen base model come from different model families.
Replacing the Qwen3-1.7B expert with Llama3.2-3B \citep{grattafiori2024llama} while keeping the base model fixed as Qwen3-8B remains competitive on both domains, achieving 69.46 on Materials and 63.97 on Adjuvant.
These results indicate that GAG is not restricted to same-family knowledge transfer, and that its latent interface can support effective specialist injection across heterogeneous expert/backbone pairings.

\begin{table}[H]
\centering
\small
\setlength{\tabcolsep}{5pt}
\renewcommand{\arraystretch}{1.10}
\caption{\textbf{Cross-family transferability of GAG.}
We replace the Qwen3-1.7B domain expert with Llama3.2-3B while keeping the frozen base model fixed as Qwen3-8B.
All scores are BertScore on a $\times 100$ scale.
\textbf{Bold} indicates the best result and \underline{underline} indicates the second best.}
\label{tab:cross-family-transfer}
\resizebox{\columnwidth}{!}{%
\begin{tabular}{@{}lccc@{}}
\toprule
\textbf{Configuration} 
& \textbf{Materials} 
& \textbf{Adjuvant} \\
& \textbf{($\times 100$)} $\uparrow$ 
& \textbf{($\times 100$)} $\uparrow$ \\
\midrule
GAG (Qwen3-1.7B expert + Qwen3-8B base)   & \underline{69.11} & \textbf{64.28} \\
GAG (Llama3.2-3B expert + Qwen3-8B base)  & \textbf{69.46} & \underline{63.97} \\
\bottomrule
\end{tabular}%
}
\end{table}


\section{More Interesting Cases}
\label{app:more-interesting-cases}

\begin{figure*}[b]
  \centering
  \includegraphics[width=\textwidth]{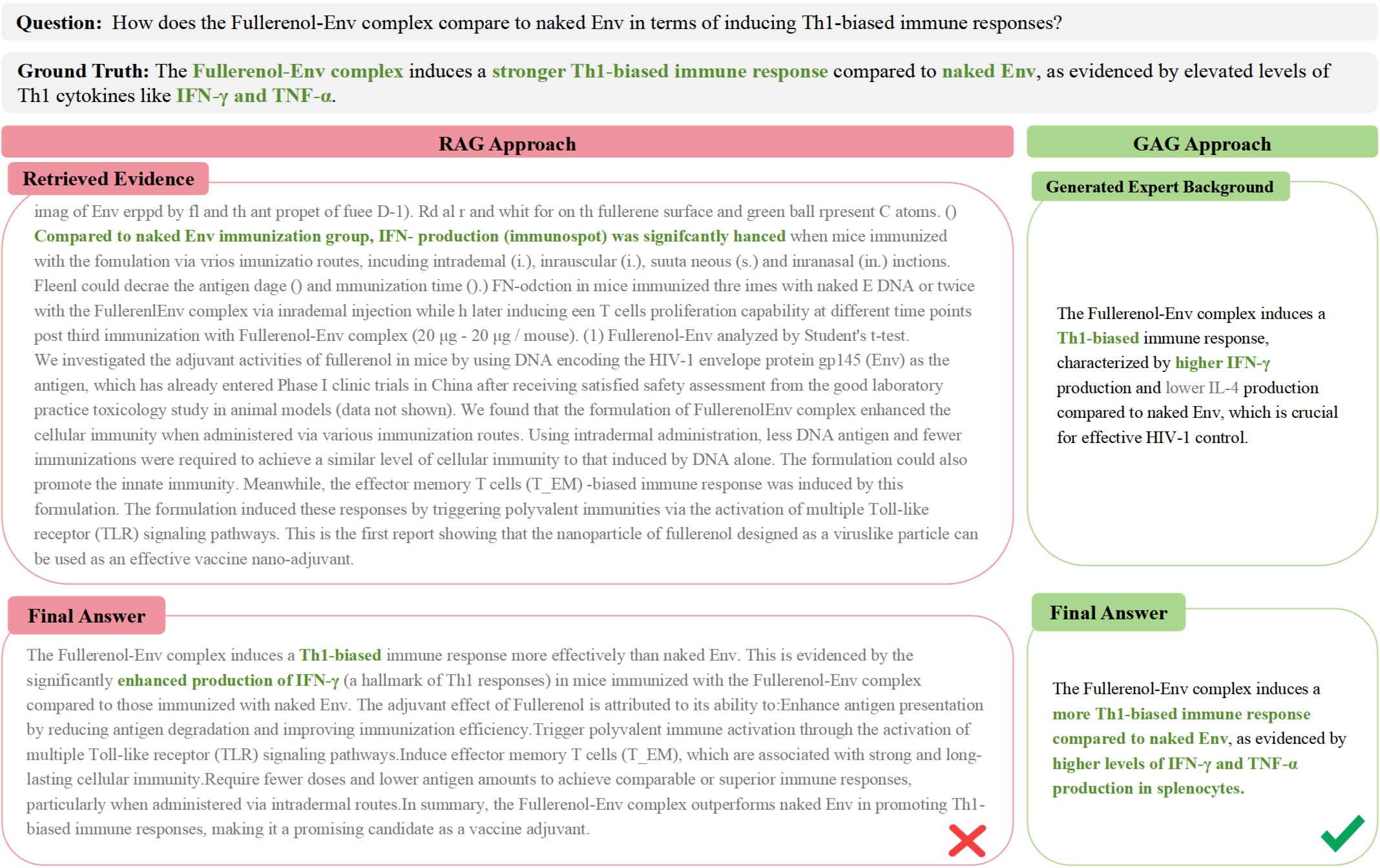}
  \caption{\textbf{Case 1 (Adjuvant): Robust key-factor coverage under noisy and fragmented retrieval.}
  RAG retrieves a corrupted and noisy snippet, which weakens evidence completeness for reference-critical cytokine signals, while GAG yields a more faithful Th1 comparison via constant-budget latent-memory injection.
  Green highlights indicate reference-critical factors.}
  \label{fig:more-case1}
\end{figure*}

\begin{figure*}[t]
  \centering
  \includegraphics[width=\textwidth]{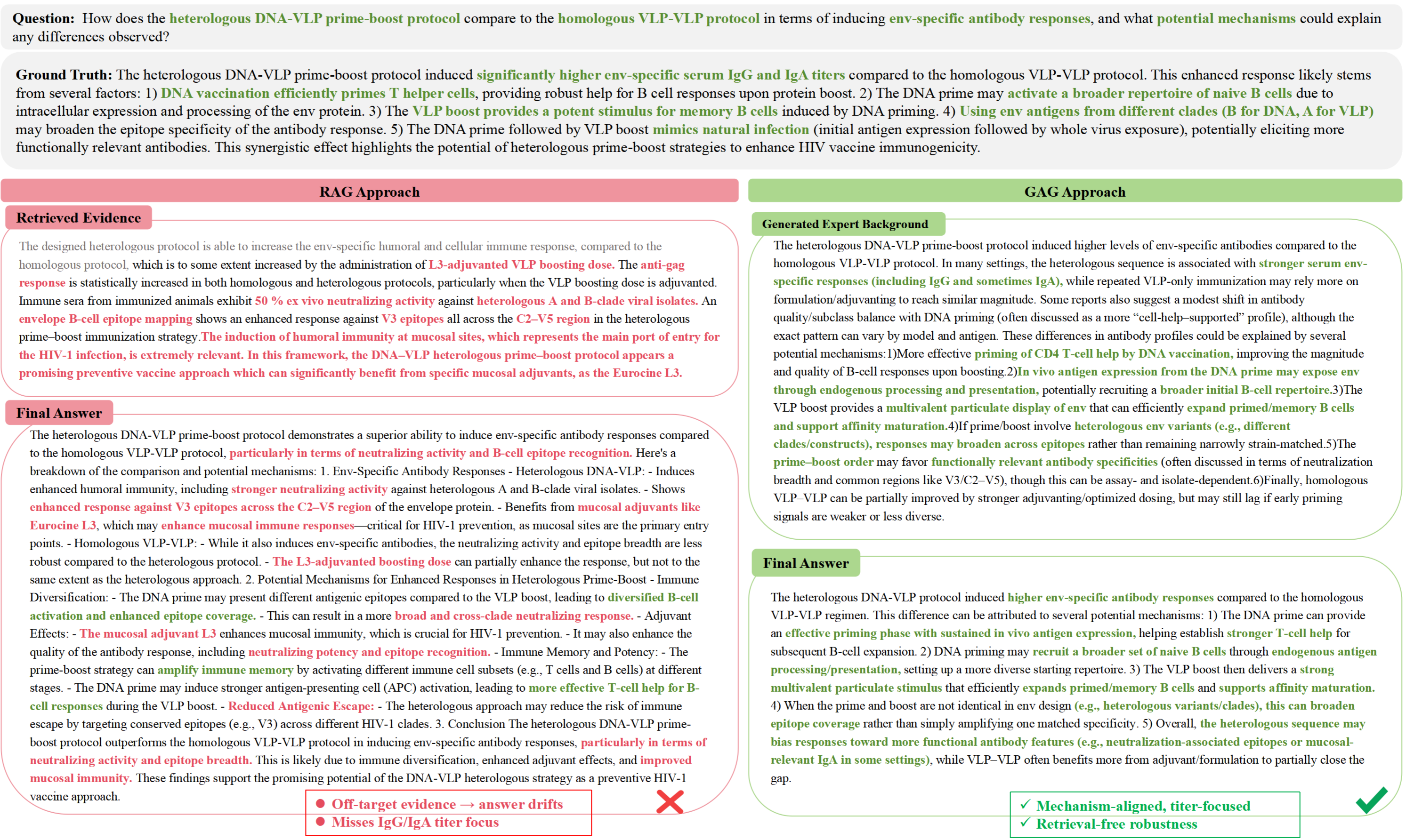}
  \caption{\textbf{Case 2 (Adjuvant): Titer-critical QA under retrieval drift.}
  The question targets env-specific serum IgG/IgA titers and their mechanistic drivers.
  RAG retrieval over-emphasizes Eurocine L3/mucosal framing, anti-gag signals, and neutralization/epitope-mapping anecdotes, inducing a neutralization-centric answer that drifts from the requested titer comparison.
  GAG instead yields a titer-aligned, mechanism-grounded explanation (DNA priming for CD4 help and broadened naive B-cell recruitment; VLP boosting for strong memory B-cell expansion/affinity maturation; heterologous clades broadening epitope coverage) via fixed-budget latent-memory injection.}
  \label{fig:more-case2}
\end{figure*}

\begin{figure*}[t]
  \centering
  \includegraphics[width=\textwidth]{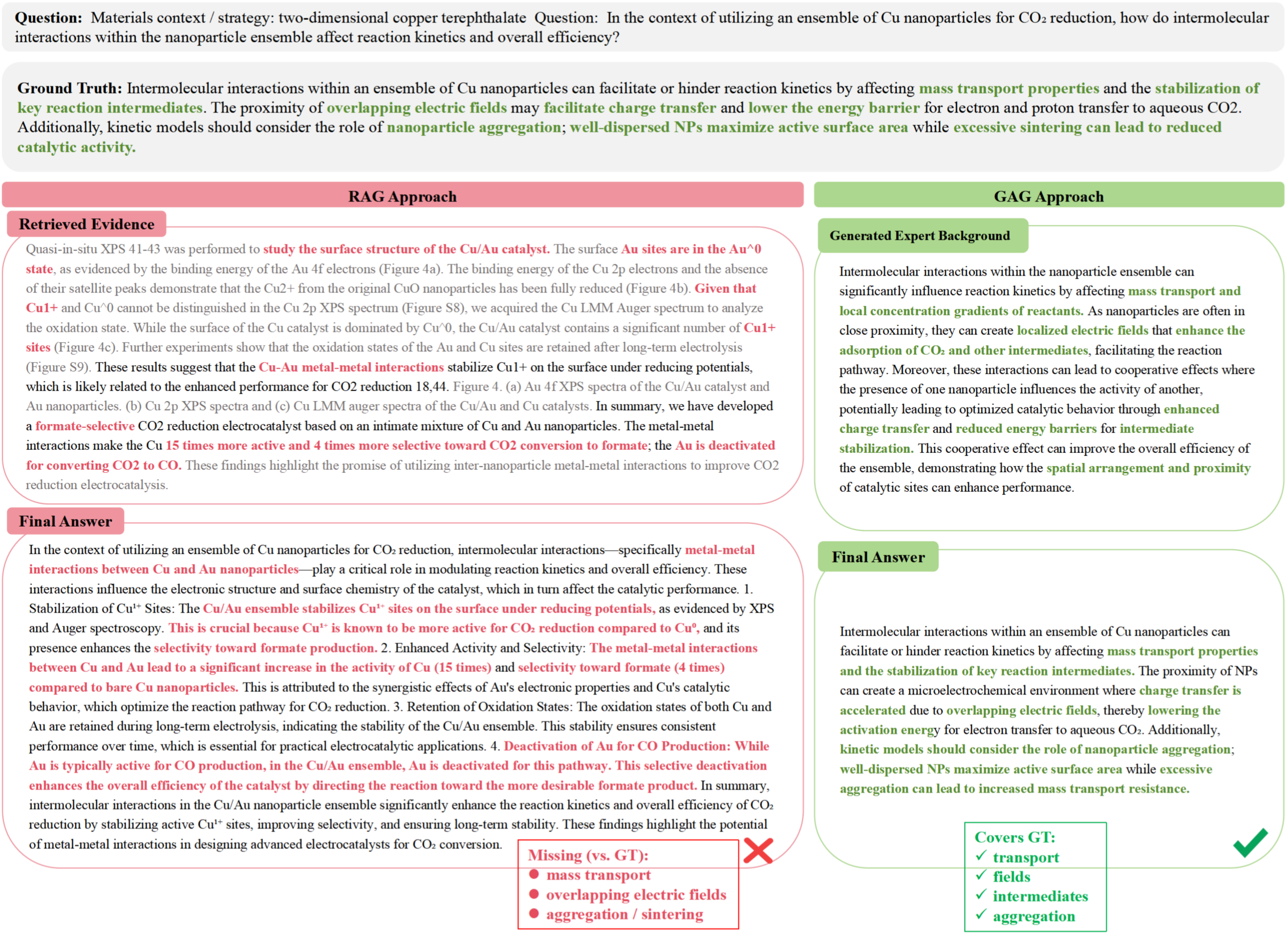}
  \caption{\textbf{Case 3 (Materials): Mechanism-level synthesis under wrong-entity retrieval.}
  RAG retrieval centers on a Cu/Au-XPS characterization narrative, inducing an off-target answer that misses the reference mechanisms (mass transport, overlapping fields, intermediate stabilization, and aggregation effects).
  GAG recovers the intended ensemble-interaction mechanisms via latent-memory injection and explicitly covers the reference factors.}
  \label{fig:more-case3}
\end{figure*}

\begin{figure*}[t]
  \centering
  \includegraphics[width=\textwidth]{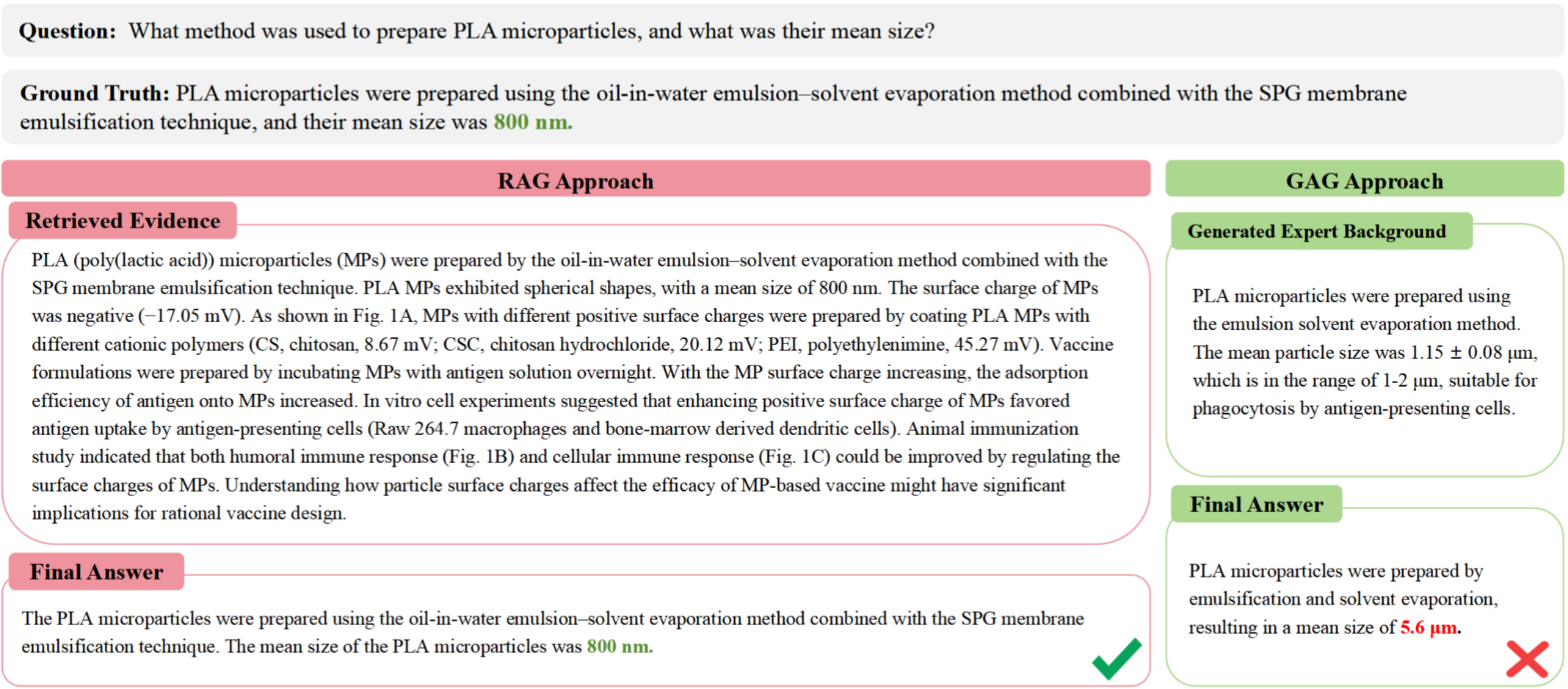}
  \caption{\textbf{Case 4 (Adjuvant; error analysis): Exact numeric fidelity vs.\ knowledge transfer.}
  RAG can precisely copy an explicitly stated mean size from retrieved text, while GAG correctly captures the preparation procedure but exhibits a minor nm/$\mu$m scale slip on the numeric value.
  This suggests that lightweight numeric normalization or verification can complement constant-budget latent-memory injection when exact numbers dominate the evaluation.}
  \label{fig:more-case4}
\end{figure*}

\textbf{Note on visualization.}
In Figures~\ref{fig:more-case1}--\ref{fig:more-case4}, the ``Generated Expert Background'' on the GAG side is shown only as an analysis-time probe for interpretability.
In the actual GAG pipeline, the domain expert model ($\mathrm{LLM}_{\text{domain},i}$) does \textbf{not} expose any such text to users, and the frozen base model does \textbf{not} consume expert-generated text directly.
Instead, GAG compresses question-conditioned multi-layer hidden states from $\mathrm{LLM}_{\text{domain},i}$ into \textbf{multi-slot latent memories}, performs cross-layer fusion, aligns them to $\mathrm{LLM}_{\text{base}}$ through a gated residual projector, and injects them through a fixed number of reserved special tokens.
Thus, the user-visible interface remains constant-budget and retrieval-free.

\textbf{Legend (highlight colors).}
Across Figures~\ref{fig:more-case1}--\ref{fig:more-case4}, \textcolor{green}{green} highlights denote ground-truth-critical key factors, \textcolor{red}{red} highlights mark off-target, mismatched, or misleading retrieved details that can derail RAG, and \textcolor{gray}{gray} text indicates irrelevant or noisy content that is not required by the reference answer.

\textbf{Case 1 (Adjuvant): noisy retrieval $\rightarrow$ incomplete evidence grounding.}
As shown in Figure~\ref{fig:more-case1}, this case illustrates a practical brittleness of RAG in private scientific corpora: the retrieved top passage is partially corrupted and fragmented, so the base model is forced to answer under incomplete and unstable evidence support.
Even when the RAG answer captures the coarse direction (Th1 bias), retrieval noise can suppress explicit coverage of all reference-critical markers.
GAG avoids this failure mode by decoupling domain knowledge transfer from snippet quality: the injected latent memories provide a more holistic specialist prior in $\mathrm{LLM}_{\text{base}}$'s representational space, enabling more reliable coverage of the key Th1 evidence emphasized by the ground truth.

\textbf{Case 2 (Adjuvant): retrieval drift $\rightarrow$ objective misalignment.}
As shown in Figure~\ref{fig:more-case2}, this example exposes a high-stakes RAG failure mode in private scientific corpora: objective-misaligned evidence can be topically relevant yet steer generation toward the wrong criterion.
The ground truth is explicitly titer-based (higher env-specific IgG/IgA under DNA--VLP than VLP--VLP) and attributes the gap to a coherent prime--boost mechanism.
However, retrieved snippets foreground adjuvant and mucosal details together with neutralization-related evidence (plus off-target readouts such as anti-gag), so the base model over-focuses on epitope breadth and neutralization narratives and under-serves the titer-focused comparison the question demands.
GAG avoids this drift by replacing snippet-level evidence serialization with a representation-level expert prior: the injected latent memories encode the causal chain needed for the titer claim, delivering higher intent fidelity under a constant knowledge budget.

\textbf{Case 3 (Materials): wrong-entity retrieval $\rightarrow$ mechanism collapse.}
As shown in Figure~\ref{fig:more-case3}, mechanism questions are particularly vulnerable to RAG's entity mismatch: when retrieval locks onto an adjacent but different experimental setup (here, Cu/Au catalyst characterization), the base model is steered into an evidence-consistent yet question-inconsistent explanation.
Consequently, RAG shifts to a Cu/Au-specific story and fails to cover the ground-truth mechanism checklist (transport effects, overlapping-field charge transfer, intermediate stabilization, and aggregation/sintering tradeoffs).
GAG mitigates this by injecting a domain-conditioned representation that is not tied to a single retrieved entity or paper chunk, allowing $\mathrm{LLM}_{\text{base}}$ to synthesize the intended cross-concept mechanism and maintain high-level faithfulness under retrieval mismatch.

\textbf{Case 4 (Adjuvant; error analysis): a minor numeric-scale slip.}
As shown in Figure~\ref{fig:more-case4}, this error case reflects a common pattern in scientific QA: retrieval can directly surface and copy exact numeric details when they appear verbatim in the retrieved span.
GAG still provides strong procedural correctness (the preparation method aligns at a high level), but shows a small unit or scale slip (nm vs.\ $\mu$m) on the mean size.
In practice, this is a lightweight edge case: when exact numeric fidelity is paramount, simple post-hoc numeric or unit normalization can be layered on top of the injected expert signal without changing the core constant-budget design.

\end{document}